\documentclass[10pt,twocolumn,letterpaper]{article}

\usepackage[pagenumbers]{cvpr} %

\usepackage[dvipsnames]{xcolor}

\usepackage[accsupp]{axessibility} %

\newcommand{\ours}{Sat2Scene\xspace}

\newcommand{\boldparagraph}[1]{\vspace{0.01em}\noindent{\bf #1}}

\newcommand{\matrixfont}[1]{\mathbf{#1}}
\newcommand{\vectorfont}[1]{\mathbf{#1}}
\newcommand{\setfont}[1]{\mathbf{#1}}
\newcommand{\networkfont}[1]{\boldsymbol{\mathcal{#1}}}

\usepackage{multirow}
\usepackage{makecell}

\newcommand{\VIDEO}{ Please use \textbf{Adobe Reader} / \textbf{KDE Okular} to see \textbf{animations}.}
\newcommand{\VIDEOREMINDER}{ Use the scroll to drag frames while the mouse is over the video. Leave the current page and back to replay.}

\usepackage[bigfiles]{pdfbase}
\ExplSyntaxOn
\NewDocumentCommand\embedvideo{smm}{
  \group_begin:
  \leavevmode
  \tl_if_exist:cTF{file_\file_mdfive_hash:n{#3}}{
    \tl_set_eq:Nc\video{file_\file_mdfive_hash:n{#3}}
  }{
    \IfFileExists{#3}{}{\GenericError{}{File~`#3'~not~found}{}{}}
    \pbs_pdfobj:nnn{}{fstream}{{}{#3}}
    \pbs_pdfobj:nnn{}{dict}{
      /Type/Filespec/F~(#3)/UF~(#3)
      /EF~<</F~\pbs_pdflastobj:>>
    }
    \tl_set:Nx\video{\pbs_pdflastobj:}
    \tl_gset_eq:cN{file_\file_mdfive_hash:n{#3}}\video
  }
  \pbs_pdfobj:nnn{}{dict}{
    /Type/RichMediaInstance/Subtype/Video
    /Asset~\video
    /Params~<</FlashVars (
    )>>
  }
  \pbs_pdfobj:nnn{}{dict}{
    /Type/RichMediaConfiguration/Subtype/Video
    /Instances~[\pbs_pdflastobj:]
  }
  \pbs_pdfobj:nnn{}{dict}{
    /Type/RichMediaContent
    /Assets~<<
      /Names~[(#3)~\video]
    >>
    /Configurations~[\pbs_pdflastobj:]
  }
  \tl_set:Nx\rmcontent{\pbs_pdflastobj:}
  \pbs_pdfobj:nnn{}{dict}{
    /Activation~<<
      /Condition/\IfBooleanTF{#1}{PV}{XA}
      /Presentation~<</Style/Embedded>>
    >>
    /Deactivation~<</Condition/PI>>
  }
  \hbox_set:Nn\l_tmpa_box{#2}
  \tl_set:Nx\l_box_wd_tl{\dim_use:N\box_wd:N\l_tmpa_box}
  \tl_set:Nx\l_box_ht_tl{\dim_use:N\box_ht:N\l_tmpa_box}
  \tl_set:Nx\l_box_dp_tl{\dim_use:N\box_dp:N\l_tmpa_box}
  \pbs_pdfxform:nnnnn{1}{1}{}{}{\l_tmpa_box}
  \pbs_pdfannot:nnnn{\l_box_wd_tl}{\l_box_ht_tl}{\l_box_dp_tl}{
    /Subtype/RichMedia
    /BS~<</W~0/S/S>>
    /Contents~(embedded~video~file:#3)
    /NM~(rma:#3)
    /AP~<</N~\pbs_pdflastxform:>>
    /RichMediaSettings~\pbs_pdflastobj:
    /RichMediaContent~\rmcontent
  }
  \phantom{#2}
  \group_end:
}
\ExplSyntaxOff

\definecolor{cvprblue}{rgb}{0.21,0.49,0.74}
\usepackage[pagebackref,breaklinks,colorlinks,citecolor=cvprblue]{hyperref}

\def\paperID{4065} %
\def\confName{CVPR}
\def\confYear{2024}

\title{\ours: 3D Urban Scene Generation from Satellite Images with Diffusion}

\author{
Zuoyue Li$^\text{1}$
\hspace{1em}
Zhenqiang Li$^\text{2}$
\hspace{1em}
Zhaopeng Cui$^\text{3}$
\hspace{1em}
Marc Pollefeys$^\text{1,4}$
\hspace{1em}
Martin R. Oswald$^\text{1,5}$
\\
$^\text{1}$ETH Z\"urich
\hspace{1em}
$^\text{2}$The University of Tokyo
\hspace{1em}
$^\text{3}$Zhejiang University
\hspace{1em}
\\
$^\text{4}$Microsoft
\hspace{1em}
$^\text{5}$University of Amsterdam
}

\begin{document}

\newcolumntype{L}[1]{>{\raggedright\let\newline\\\arraybackslash\hspace{0pt}}m{#1}}
\newcolumntype{C}[1]{>{\centering\let\newline\\\arraybackslash\hspace{0pt}}m{#1}}
\newcolumntype{R}[1]{>{\raggedleft\let\newline\\\arraybackslash\hspace{0pt}}m{#1}}

\maketitle

\begin{abstract}

Directly generating scenes from satellite imagery offers exciting possibilities for integration into applications like games and map services.
However, challenges arise from significant view changes and scene scale.
Previous efforts mainly focused on image or video generation, lacking exploration into the adaptability of scene generation for arbitrary views.
Existing 3D generation works either operate at the object level or are difficult to utilize the geometry obtained from satellite imagery.
To overcome these limitations, we propose a novel architecture for direct 3D scene generation by introducing diffusion models into 3D sparse representations and combining them with neural rendering techniques.
Specifically, our approach generates texture colors at the point level for a given geometry using a 3D diffusion model first, which is then transformed into a scene representation in a feed-forward manner.
The representation can be utilized to render arbitrary views which would excel in both single-frame quality and inter-frame consistency.
Experiments in two city-scale datasets show that our model demonstrates proficiency in generating photo-realistic street-view image sequences and cross-view urban scenes from satellite imagery.

\end{abstract}
    
\section{Introduction}
\label{sec:intro}

The direct generation of scenes from satellite imagery sparks exciting possibilities for seamlessly integrating diverse real-life environments into games, films, and map services.
However, this task remains challenging due to the significant view differences between satellite images and street-view images on the ground, coupled with the expansive scale of the scenes under consideration.
Prior efforts primarily focused on image or video generation~\cite{sat2img,sat2str,sat2density,sat2vid}, with relatively limited exploration into scene generation.
The salient advantage of scenes resides in their adaptability for rendering images from arbitrary views, which is a feature particularly beneficial for the aforementioned practical applications.
Conventional cross-view synthesis works \cite{sat2img,sat2str,sat2density} tend to generate street-view images from the geometry estimated based on the satellite height map.
Their limitations arise from the individual generation through 2D GANs, leading to the resulting images holding a noticeable lack of inter-frame consistency.
Sat2Vid~\cite{sat2vid} further advances the generation of street-view images with multi-view consistency by explicitly building the 2D-3D correspondence map between pixels in frames and point clouds. 
Nevertheless, the 3D point cloud is constructed by back-projecting pixels in all 2D frames, restricting its ability to synthesize video along predefined trajectories rather than arbitrary views.
Furthermore, memory usage significantly increases with growing trajectory length, imposing constraints on the permissible video length.

\begin{figure}[!t]
    \centering
    \scriptsize
    \newcommand{\sz}{0.48\textwidth} %

    \begin{tabular}{C{0.135\textwidth}C{0.14\textwidth}C{0.135\textwidth}}
    Satellite w/ trajectory & Generated ground-view & Generated bird-view \\
    \end{tabular}
    \embedvideo*{\includegraphics[width=\sz]{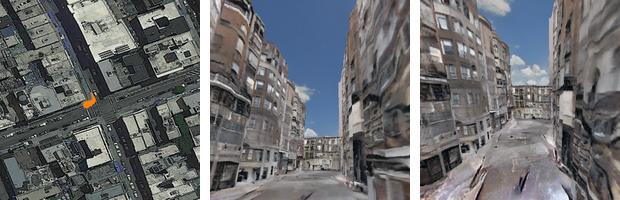}}{media/teaser/teaser_1_b800k.mp4}
    \embedvideo*{\includegraphics[width=\sz]{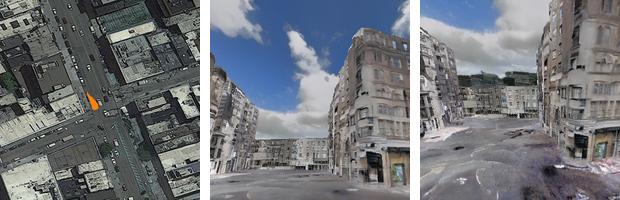}}{media/teaser/teaser_2_b800k.mp4}
    
    \caption{\textbf{Urban scenes generated by Sat2Scene.}
    From a single satellite image covering urban streets, Sat2Scene is able to generate videos with photorealistic and consistent textures across different views.\VIDEO\VIDEOREMINDER
    }
    \label{fig:teaser}
\end{figure}

In recent years, notable advancements~\cite{instantngp,nsvf,pointnerf} in scene representations have been observed, particularly in the realm of novel view synthesis since the introduction of NeRF~\cite{nerf}. 
Inspired by this, InfiniCity~\cite{infinicity} relies on the pseudo ground truth generated by the 2D network for per-scene training through the optimization of the radiance field.
However, it exhibits limitations in flexibility and is contingent on the performance of the pre-trained 2D network.%

On the other hand, diffusion models have brought generative models to a new era due to their better performance and generative ability than GANs~\cite{diffmodelbeatgan}.
In the realm of 3D generation, the majority of these models were introduced for generating 3D objects~\cite{diffrf,uvdiff} based on dense voxels.
However, large-scale scenes occupy an excessive number of voxels, demanding lots of memory and restricting the scalability of these models for outdoor scene generation.

To address the aforementioned issues, we propose a novel architecture that harnesses the potential generative ability of the diffusion model coupled with the photorealistic rendering effect provided by neural rendering.
This architecture directly generates the scene representation tightly associated with the geometry at the point level and produces consistent images from any view by assembling the features in 3D space through a physical-based renderer.
Specifically, we accomplish this by implementing a diffusion model based on 3D sparse representations to derive point-wise color through direct denoising on the noised point clouds.
We subsequently extract point-anchored features from the initially colorized point cloud and feed them to the neural rendering network with point-wise feature aggregation.
This allows us to overcome the resolution limitations of voxel-based representations while ensuring efficient memory usage for handling large-scale outdoor scenes, and to generate images from arbitrary views in the scene while simultaneously maintaining consistency.
The proposed architecture surpasses state-of-the-art generation models on the HoliCity~\cite{holicity} dataset in terms of the overall quality of the rendered video and inter-frame consistency.
Additionally, the trained model demonstrates superior generalization to another city-scale dataset OmniCity~\cite{omnicity} further containing satellite images.
\cref{fig:teaser} presents rendered videos of our method inferred on two scenes from the satellite images, which demonstrates both photorealism and robust consistency across views.

\noindent The \textbf{contributions} of this paper are three-fold:
\begin{itemize}
\item We present a novel diffusion-based framework \ours for direct 3D scene generation, which is able to generate 3D urban scenes just from satellite images.
\item To ensure consistent image generation from any view, a novel diffusion model with sparse representations is proposed to generate scene features tightly associated with the geometry directly in 3D space. To the best of our knowledge, we are the first to combine diffusion models with 3D sparse representations.
\item Our model demonstrates the capability to produce photo-realistic street-view image sequences with robust temporal consistency. Superior performance compared to state-of-the-art baselines validates this ability.
The model exhibits proficiency when employed for cross-view urban scene generation from satellite images.
\end{itemize}

\section{Related work}
\label{sec:relat}

Our work is at the cross-section of 3D generative models and neural rendering.
In the following, we discuss related works according to their underlying categories.

\boldparagraph{Diffusion models} have gradually become the dominant generative models in recent years since DDPM~\cite{ddpm}. 
These score-based generative models typically hold more stable training, do not require discriminators, and have been proven to achieve better quality than GANs~\cite{diffmodelbeatgan}.
Advanced techniques such as DDIM~\cite{ddim} and its follow-up \cite{improvedddpm} accelerate the sampling process while maintaining generation quality or better time scheduling during training.
Latent diffusion models (LDMs)~\cite{ldm} further increase the denoising stability by operating in a compressed latent space to reduce the computational cost and memory footprint, making the synthesis of high-resolution images feasible on consumer devices.
While image quality is constantly improving, how to utilize diffusion models to ensure multi-view consistency for multiple images generated from the same scene is currently yet less studied, which is crucial for scene or video generation.
The concurrent work of this paper, MVDiffusion~\cite{mvdiffusion}, enables holistic multi-view image generation building by leveraging LDMs and ControlNet~\cite{controlnet}, however, its cross-frame consistency remains suboptimal.

\boldparagraph{3D diffusion models}, once predominantly concentrated on geometry generation~\cite{pointdiff, hyperdiff}, have evolved to encompass texture generation.
This expansion achieves better multi-view consistency particularly when coupled with neural differentiable rendering techniques.
DiffRF~\cite{diffrf} synthesizes a radiance field based on diffusion probabilistic models performed in 3D space.
The direct denoising process on the 3D feature space of the radiance field ensures the consistency of the rendered results from different views.
Point-UV Diffusion \cite{uvdiff} employs a two-level diffusion structure to produce coarse-to-fine texture for mesh. The initial coarse texture is produced on the 3D dense voxel grids of the mesh and is subsequently refined by a secondary diffusion model in 2D space. 
However, due to the memory-intensive nature of the dense volumetric representation, scalability to large scenes is constrained.
NF-LDM~\cite{nfldm} solves the issue by compressing the dense scene grid into three disentangled compact representations using an auto-encoder, and training separate diffusion models for each representation. 
The sampled compact representations are then fed into the decoder section of the auto-encoder to reconstruct the original scene grid, which is subsequently employed for rendering arbitrary views.
However, it might not be useful for generating content on a specific geometry, which requires further post-optimization procedures.

\boldparagraph{3D-aware generative models} that do not rely on diffusion models have been extensively researched with various scene representations.
PixelNeRF~\cite{pixelnerf} extends NeRF~\cite{nerf} in a feed-forward manner without test-time optimization.
The sampling points in the rendering ray aggregate features from given views, which are used for color and density estimation. 
However, due to the less accurate point correspondence between views, the rendered results often exhibit blurriness.
Alternatively, some methods integrate GAN settings with neural rendering techniques. 
They commonly employ intermediate explicit representations associated with simplified space structures or sparse grids to enhance memory efficiency and scalability for scenes of varying sizes.
For example, GSN~\cite{gsn} generates a 2D floor plan as the intermediate explicit representation for unconditional indoor living scene generation.
Similarly, EG3D~\cite{eg3d} adopts a tri-plane representation, contributing to both the efficiency and effectiveness of the generation process.
CompNVS~\cite{compnvs} tries to solve the 3D inpainting or completion task via conditional generation based on NSVF~\cite{nsvf} representation using a sparse grid.
GANcraft~\cite{gancraft} is designed for large-scale scene generation given sparse geometry with semantic information. 
The approach leverages a pre-trained SPADE model~\cite{SPADE} to generate pseudo ground-truth images, which are then used to perform a NeRF-like optimization with a style controller.
Notably, GANcraft's performance is heavily dependent on the image quality of the 2D generator and necessitates per-scene training, which is non-feed-forward.

\boldparagraph{Satellite-to-ground synthesis} is a very typical application of cross-view generation, which focuses on synthesizing images from a distinctly different perspective of a provided top-view satellite image.
S2G~\cite{sat2img} is the first work in this area, considering geometry consistency in satellite-to-ground synthesis. 
The approach predicts a density voxel grid from the satellite height map and transforms it to ground depth and semantics panorama, followed by a 2D generative model to obtain the street view.
Its follow-up Sat2Vid~\cite{sat2vid} further addresses the temporal inconsistency challenge when generalizing S2G to video synthesis, via maintaining a correspondence map between 3D points and 2D frame pixels.
The generator performs directly in 3D space, facilitating the natural maintenance of multi-view consistency in the rendered frames, showing better plausibility than 2D video-to-video translation advances Vid2Vid~\cite{vid2vid} and WC-Vid2Vid~\cite{wcvid2vid}.
Nevertheless, the point cloud is pre-defined following the rendering poses hence the size of the point cloud would explode with the length of the video sequence increasing.
Recently, InfiniCity~\cite{infinicity} implements the satellite-to-ground generation on a city-wide scale.
However, since it adopts GANcraft~\cite{gancraft} as the generative model, the method still requires per-scene training and test-time optimization.
Sat2Density~\cite{sat2density} places emphasis on top-view density prediction without depth supervision.
It achieves this by leveraging the view relationship between satellite and ground, and integrating neural rendering techniques to enhance synthesis quality. 
But due to this, the resulting videos lack good temporal consistency.

\section{Method}
\label{sec:method}

\begin{figure*}[ht]
    \centering
    \includegraphics[width=\linewidth]{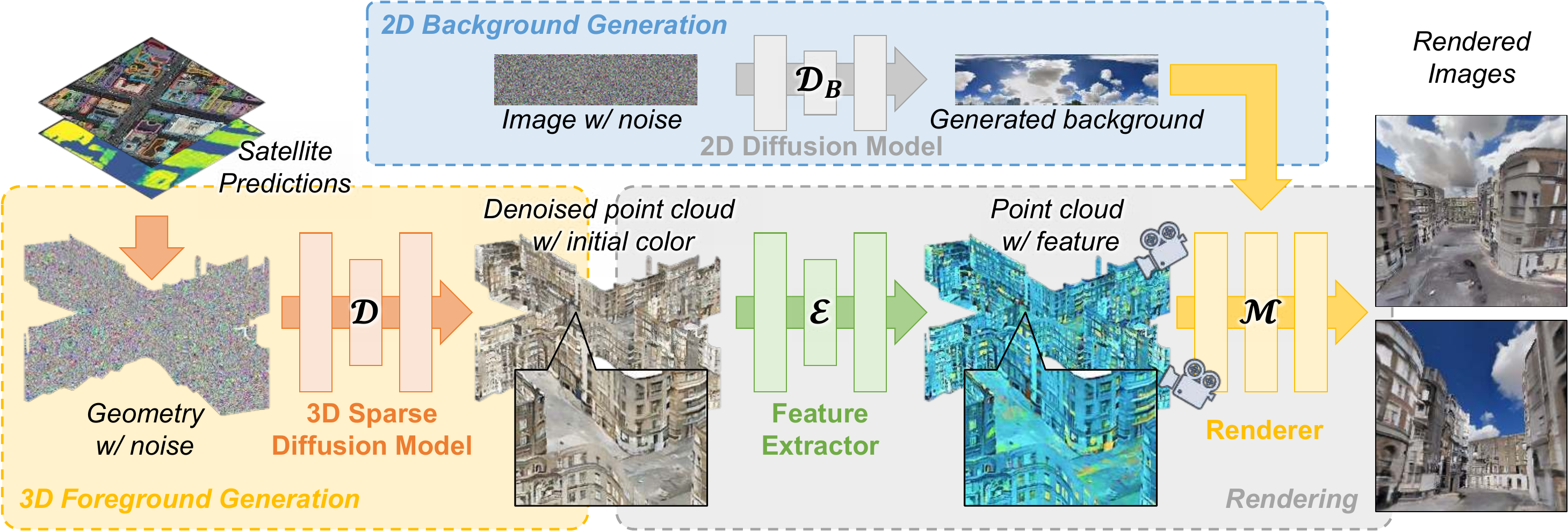}
    \caption{\textbf{Pipeline overview of our method.} Three steps compose the full pipeline to generate the scene representation and render street views based on satellite-inferred geometries. The \textbf{generation step} initiates colors for the foreground point cloud by using a 3D diffusion model with sparse convolutions, as well as synthesizing the background panorama with a 2D diffusion model. The scene features tightly anchored with the point cloud are extracted at the \textbf{feature extraction step}. The final \textbf{rendering step} produces images from arbitrary views through neural rendering.}
    \label{fig:pipeline}
\end{figure*}

Our method disentangles the entire scene into two components: the foreground, comprising buildings and roads, and the background sky. 
We represent the foreground as a point cloud, denoted as $(\matrixfont{P}, \matrixfont{F})$, where $\matrixfont{P} \in \mathbb{R}^{N \times 3}$ is the world coordinates of points in 3D space, $\matrixfont{F} \in \mathbb{R}^{N \times D}$ represents per-point-associated feature map with $D$ channels, and $N$ is the number of points. 
The background is modeled as a dome at an infinite distance and represented by a 2D panorama image $\matrixfont{B} \in \mathbb{R}^{H_B \times W_B \times 3}$ with height $H_B$ and width $W_B$. 
Our pipeline is specifically designed to generate these representations for synthesizing texture information of a given input geometry $\matrixfont{P}$.
As illustrated in the \cref{fig:pipeline}, the proposed pipeline mainly comprises three steps. 
The generation phase initiates color scenes by colorizing the foreground point clouds with \textbf{a diffusion model in a 3D sparse setting} and synthesizing the background skies using a 2D diffusion model (\cref{sec:gen}).
Then the foreground features are extracted from the initial color scene with a 3D encoder in a feed-forward manner.
Finally, the neural renderer produces the color frames for the given poses (\cref{sec:render}).

\subsection{Generation}
\label{sec:gen}

The generation phase creates initial textures, producing a per-point color map $\matrixfont{C} \in [0, 1]^{N \times 3}$ on the normalized RGB space for the fixed geometry $\matrixfont{P}$. 
We have generalized diffusion model techniques~\cite{ddpm,ddim,improvedddpm} to the 3D space with a sparse setting. Specifically, the 3D denoising network $\networkfont{D}$ utilizes 3D sparse convolutional layers, which operate exclusively on the occupied areas (building facades and road surfaces). 
The loss function of $\networkfont{D}$ is defined as
\begin{equation}
\label{eq:loss_foreground}
L_F = \mathbb{E}_{\matrixfont{C},\boldsymbol{\epsilon},t} \left[ || (\boldsymbol{\epsilon} - \networkfont{D}(\matrixfont{P}, \matrixfont{C}_t, t)) \otimes \vectorfont{m} ||_F^2 \right],
\end{equation}
where $t$ is uniformly sampled from $\{1,\ldots,T\}$ with $T$ the number of denoising steps, $\boldsymbol{\epsilon} \sim \mathcal{N}(\matrixfont{0}, \matrixfont{I})$, $\matrixfont{C}_t$ denotes the noisy version of the color map $\matrixfont{C}$ at time step $t$, $\vectorfont{m} \in [0,1]^{N}$ is a mask indicating the point confidence with $\otimes$ the broadcasted Hadamard multiplication.
The background generation follows the SOTA diffusion model~\cite{ldm}, denoting $\networkfont{D}_B$, resulting in a panoramic sky image $\matrixfont{B} \in [0, 1]^{H_B \times W_B \times 3}$. 

\subsection{Rendering}
\label{sec:render}

The feature extraction step serves as a bridge between the generation and rendering stages by extracting 3D scene features for neural rendering.
Instead of fitting a radiance field during test-time optimization, the scene representation $(\matrixfont{P}, \matrixfont{F})$ is obtained in a feed-forward way. 
A 3D point cloud encoder $\networkfont{E}$ is employed to extract the features $\matrixfont{F} = \networkfont{E}(\matrixfont{P}, \matrixfont{C})$. 
The features are individually anchored to the points $\matrixfont{P}$, representing the local scene appearance and geometry information.
The rendering stage utilizes the physically-based volume rendering techniques~\cite{nerf, pointnerf} to synthesize frames with multi-view consistency.
Suppose we have sampled $M$ points along a single rendering ray with origin $\vectorfont{o}$ and direction $\vectorfont{d}$. 
$\vectorfont{x}_i$ denotes the 3D location of the $i$-th point with $i \in \{1,\ldots,M\}$ from near to far. 
At each location $\vectorfont{x}_i$, we retrieve $K$ nearest neighboring neural points around the point within a certain radius $R$ and obtain the set of indices $\setfont{K}_i = \text{KNN}(\vectorfont{x}_i, \matrixfont{P}, K, R)$. 
The feature $\vectorfont{f}_i$ for the location $\vectorfont{x}_i$ is aggregated by a linear combination of the features anchored to the points with indices in $\setfont{K}_i$, \ie, $\vectorfont{f}_i = \frac{\matrixfont{F}^\top\vectorfont{w}_i}{||\vectorfont{w}_i||_1}$, where each element of the weight vector $\vectorfont{w}_i$ is defined as $w_{ik} = \frac{1}{||\vectorfont{x}_i - \matrixfont{P}_k||_2}, \forall k \in \setfont{K}_i$ and $w_{ik} = 0, \forall k \notin \setfont{K}_i$. 
A network $\networkfont{M}$ is used to infer the color and the density of each sampling point $(\vectorfont{c}_i, \sigma_i) = \networkfont{M}(\vectorfont{f}_i, \vectorfont{d})$.
The background color $\vectorfont{c}_0$ of the rendering ray is obtained by querying the 2D panoramic map $\matrixfont{B}$ with bilinear interpolation using the latitude and longitude coordinates computed from the ray direction $\vectorfont{d}$, \ie, $\vectorfont{c}_0 = \text{Interp}(\matrixfont{B}, \vectorfont{d})$.
The final predicted color $\hat{\vectorfont{c}}$ and depth $\hat{z}$ of the pixel are blended along the ray using volume rendering as 
\begin{equation}
\label{eq:render}
\begin{bmatrix}
\hat{\vectorfont{c}}\\
\hat{z}
\end{bmatrix}
= \sum_{i=1}^{m} \tau_{i-1}\big(1-\text{exp}(-\sigma_i\delta_i) \big)
\begin{bmatrix}
\vectorfont{c}_i\\
z_i
\end{bmatrix}
+ \tau_m
\begin{bmatrix}
\vectorfont{c}_0\\
0
\end{bmatrix},
\end{equation}
where $\tau_0 = 1$, $\tau_i = \prod_{j=1}^{i}\text{exp}(-\sigma_j\delta_j)$ denotes the accumulated transmittance along the ray, $\delta_i$ represents the distance between adjacent sampled points, and $z_i$ is the Z coordinate of the sampling point in the camera coordinate system.
The loss function $L_{\vectorfont{d}}$ of the ray with direction $\vectorfont{d}$ is defined as 
\begin{equation}
\label{eq:loss_render}
L_{\vectorfont{d}} = ||\hat{\vectorfont{c}}-\vectorfont{c}||_1 + \beta|\hat{z}-z|,
\end{equation}
where $\beta$ is a coefficient controlling the importance of depth signal during training, $\vectorfont{c}$ and $z$ are the ground-truth color and depth value of the pixel, respectively.

\subsection{Implementation details}
\label{sec:impl}

\begin{figure}[!ht]
    \centering
    \small
    \setlength{\tabcolsep}{3pt}
    \begin{tabular}{ccc}
    \includegraphics[width=0.305\columnwidth,trim={6cm 2cm 7cm 5cm},clip]{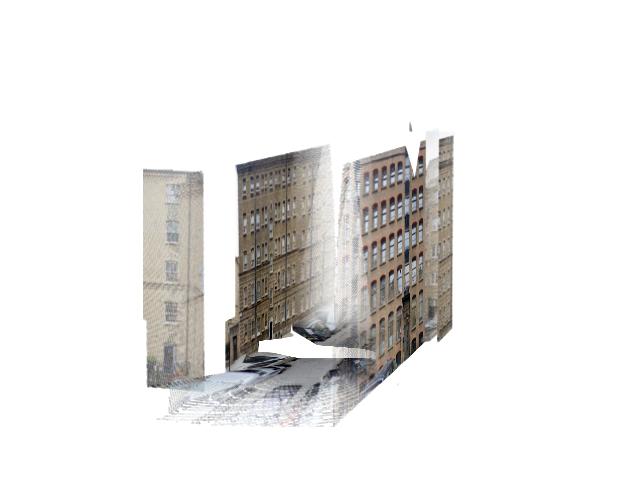} &
    \includegraphics[width=0.305\columnwidth,trim={6cm 2cm 7cm 5cm},clip]{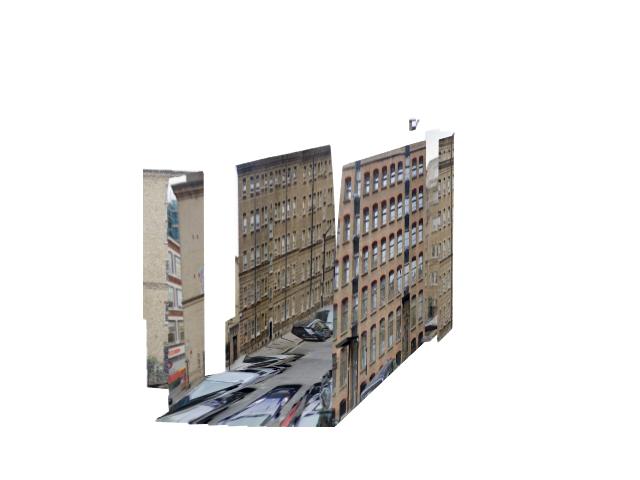} &
    \includegraphics[width=0.305\columnwidth,trim={6cm 2cm 7cm 5cm},clip]{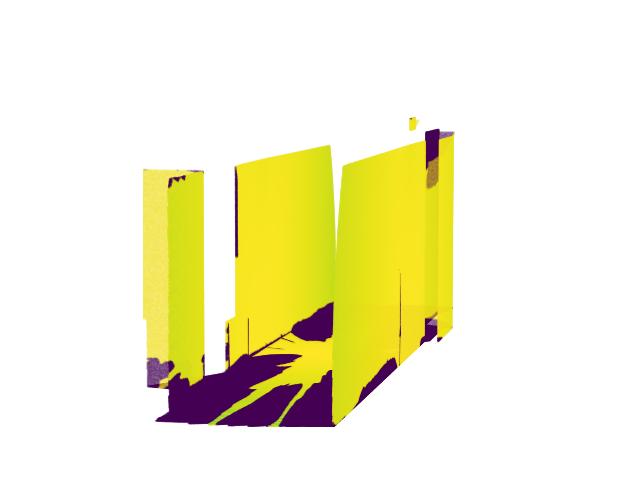} \\
    (a) w/o resampling & (b) w/ resampling & (c) point confidence
    \end{tabular}
    \caption{\textbf{Visualization of the point resampling scheme.} Yellow / purple color means high / low point confidence.}
    \vspace{-0.8em}
    \label{fig:resample}
\end{figure}

\boldparagraph{Point resampling.}
We notice that the training of the 3D sparse diffusion model is notably affected by the structure of the point cloud, particularly the balance in point density. 
The denoising process may fail when the point cloud lacks a balanced density in space, as observed in cases where the point cloud is derived from a pinhole view.
This observation emphasizes the importance of maintaining a well-distributed point cloud for effective training and denoising performance.
Therefore, the point cloud input to our pipeline is evenly sampled from the given geometry, ensuring a balanced density for improved training and generation results.
The resampling scheme adopts the Poisson disk method~\cite{poissondisk} on the geometry surfaces, where each point has approximately the same distance to the neighboring points.
\cref{fig:resample} (a) and (b) illustrate an example of the point cloud before and after resampling. (a) is fused from multiple pinhole views thus nearby points are dense and distant points are sparse, and (b) is the resampling result where the point cloud is uniformly distributed.
The colors of resampled points are interpolated and their confidence is computed based on their distance to the originally existing points. 
The confidence map $\vectorfont{m}$ is used for computing the denoising loss in \cref{eq:loss_foreground}.
For experiments using different point cloud structures, please refer to \cref{sec:ablation}.

\boldparagraph{Training.}
The entire pipeline is implemented and trained using PyTorch and Minkowski Engine~\cite{minkowski}. 
The foreground denoising network $\networkfont{D}$ is a time-conditional UNet~\cite{unet}, while the feature extractor $\networkfont{E}$ is a vanilla UNet.
Both networks adopt the provided default network architectures with 3D sparse voxel convolutions.
In the 3D diffusion model, we use $T$$=$1000, and for the rendering phase, the hyper-parameters are set as follows:  $D$$=$32, $K$$=$12, $R$$=$3m, and $\beta$$=$0.5. %
The framework runs on a single NVIDIA Tesla A100 GPU with a memory of 40GB.
The generation and rendering phases are trained separately. 
The training of the 3D diffusion model takes approximately 5 days, while the training of the neural renderer takes around 2 days, both starting from scratch.
The network voxelization size is set to 6.67cm and 5cm in the two phases respectively in order to fit the GPU memory.

\section{Experiments} \label{sec:exp}

\begin{figure*}[!ht]
    \centering
    \footnotesize

    \hspace{-0.39cm}
    \begin{tabular}{C{0.0000001\textwidth}C{0.97\textwidth}}
        \rotatebox{90}{GT} & \embedvideo*{\includegraphics[width=0.97\textwidth]{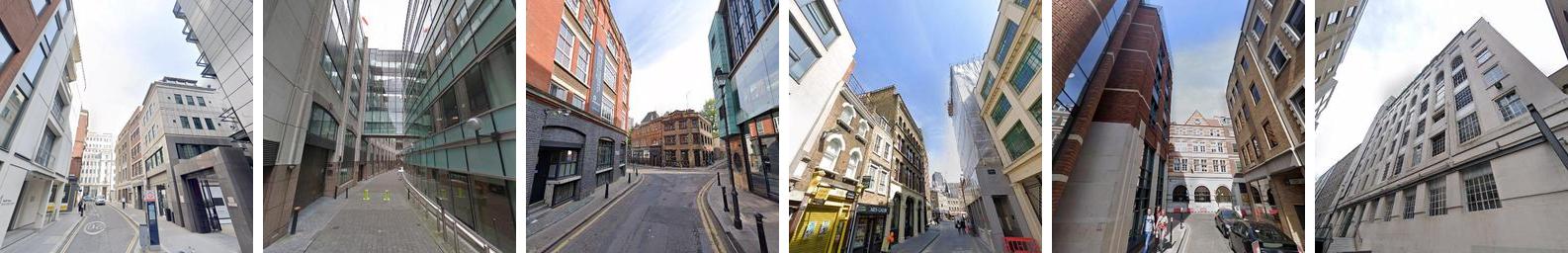}}{media/baseline_comp/GT_6res_b800k.mp4} \\
        \rotatebox{90}{Sat2Vid~\cite{sat2vid}} & \embedvideo*{\includegraphics[width=0.97\textwidth]{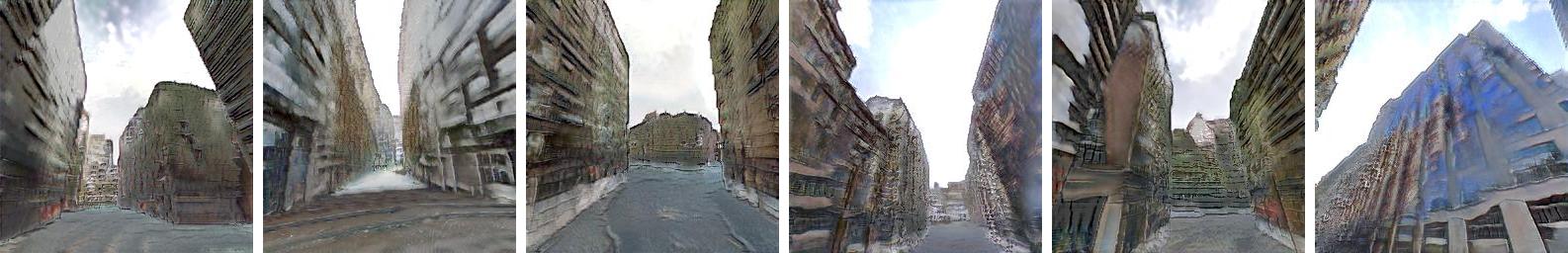}}{media/baseline_comp/Sat2Vid_6res_b10k.mp4} \\
        \rotatebox{90}{MVDiffusion~\cite{mvdiffusion}} & \embedvideo*{\includegraphics[width=0.97\textwidth]{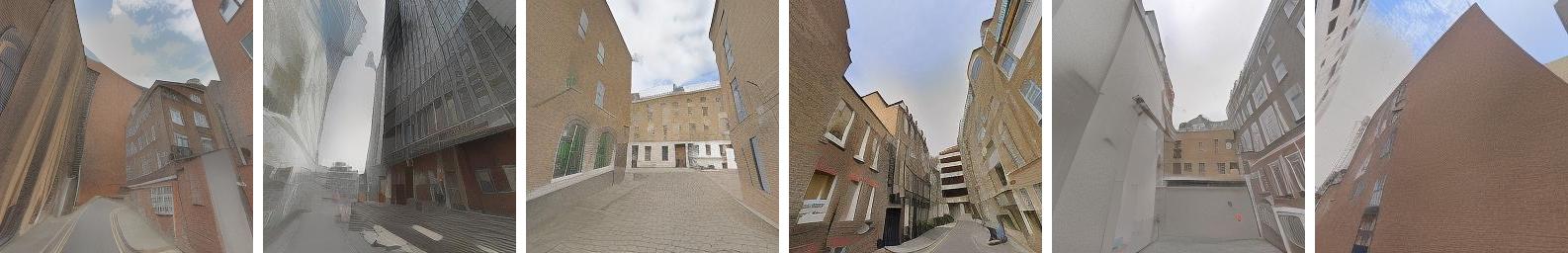}}{media/baseline_comp/MVDiff_6res_b800k.mp4} \\
        \rotatebox{90}{Ours} & \embedvideo*{\includegraphics[width=0.97\textwidth]{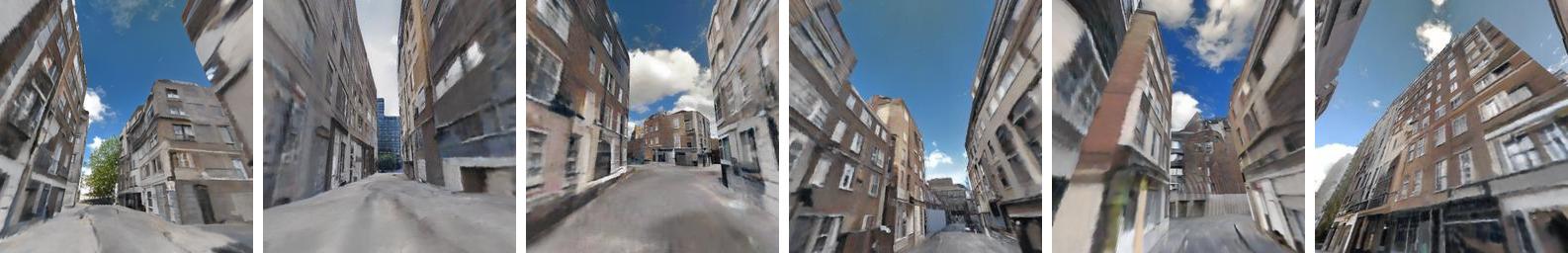}}{media/baseline_comp/Ours_6res_b800k.mp4} \\
    \end{tabular}
    
    \caption{\textbf{Qualitative baseline comparison on the HoliCity~\cite{holicity} dataset.} Our method produces higher-quality video with better temporal consistency compared with the baselines.\VIDEO}
    \label{fig:sota}
    \vspace{-0.7em}
\end{figure*}

\subsection{Configuration}
\boldparagraph{Datasets.}
We conduct our experiments on two datasets, HoliCity~\cite{holicity} and OmniCity~\cite{omnicity}.
HoliCity is a city-scale 3D dataset with rich annotation, containing 6k+ real-world urban scenes in central London.
Each scene has a corresponding high-resolution panorama image, a depth (distance) map, and 8 posed RGB-D pinhole views.
Since the CAD model is not publicly available, we obtain the scene geometry by lifting the pixels in the panorama image using the distance map. 
A scene mesh is built using the pixel connectivity in the panorama image and the point cloud is sampled on the surface using the resampling scheme described in \cref{sec:impl}.
The number of points $N$ is around one million for each scene, with a density of 400/m\textsuperscript{2}. 
The geometry for dynamic objects in the scene such as cars and pedestrians is unavailable in the dataset, thus we used the segmentation method Mask2Former~\cite{mask2former} with ViT-Adapter~\cite{vitadapter} backbone pre-trained on the Cityscapes dataset~\cite{cityscapes} to obtain the semantics of the panorama image and further set the confidence of points with mismatched semantics and geometry to zero.
\cref{fig:resample} (c) gives a visualization example of the confidence map of the points.
OmniCity is another city-scale dataset captured in New York City, further containing satellite images associated with ground-view panoramas.
However, its street-level geometry is not available, making it difficult to compare with the baselines, thus we show qualitative results for demo purposes in \cref{sec:generalization}.

\boldparagraph{Metrics.}
We evaluate our model and the baselines using the combined validation and test splits of the HoliCity dataset, containing 624 scenes and around 5k single images in total.
For quantitative evaluation, we adopt commonly used FID~\cite{fid} and KID~\cite{kid} to measure the generated image quality by comparing the distribution similarity between sets of synthesized \textbf{single} frames and real images.
We evaluate the temporal consistency across video frames via lower-level metrics PSNR, SSIM, and LPIPS~\cite{lpips}, which are only computed in the overlapping areas between neighboring frames.
Also, considering that none of the above metrics can simultaneously measure generation quality and inter-frame consistency, we further adopt FVD and KVD~\cite{fvd}, which are the video version of FID \& KID, being able to focus more on the overall quality.
They use video-classification network I3D~\cite{i3d} as the backbone, where the spatiotemporal features are utilized and the temporal relationship can be well modeled.
To overcome the insufficient number of test scenes to compute the metrics, during calculation, we pass augmented videos with each frame as the starting frame circularly, as the 8 poses of each scene in the dataset can form a loop.
This enhancement ensures that the resulting number of videos is still enough to compute those two metrics.
Additionally, we incorporated a user study of 32 people to evaluate the quality of generated videos, consisting of 30 multiple-choice questions.
For each question, the participant was asked to select a single generated video from three options (baselines and ours) that they considered more visually plausible.

\begin{table*}[ht]
    \centering
    \begin{tabular}{l|cc|cc|ccc|c}
        \toprule
        \textbf{Method / Metric} & FVD$\downarrow$ & KVD$_{\times\text{100}}$$\downarrow$ & FID$\downarrow$ & KID$_{\times\text{100}}$$\downarrow$ & PSNR$\uparrow$ & SSIM$\uparrow$ & LPIPS$\downarrow$ & User study \\
        \midrule
        Sat2Vid~\cite{sat2vid} & 37.06 & 4.03$^{\pm\text{0.05}}$ & 137.84 & 13.76$^{\pm\text{0.10}}$ & 25.25 & 0.741 & 0.252 & 2.92\% \\
        InfiniCity~\cite{infinicity} & - & - & 108.47 & 8.40$^{\pm\text{0.10}}$ & - & - & - & - \\
        MVDiffusion~\cite{mvdiffusion} & 22.79 & 2.36$^{\pm\text{0.03}}$ & \textbf{50.78} & \textbf{4.14}$^{\pm\text{0.07}}$ & 17.56 & 0.593 & 0.259 & 15.62\% \\
        \midrule
        \textbf{Ours} & \textbf{20.30} & \textbf{1.90}$^{\pm\text{0.03}}$ & 71.98 & 5.91$^{\pm\text{0.06}}$ & \textbf{31.54} & \textbf{0.956} & \textbf{0.237} & \textbf{81.46\%} \\
        \bottomrule
    \end{tabular}
    \caption{\textbf{Quantitative baseline comparison.} Sat2Scene outperforms the baselines in most of the metrics including the overall video quality and the temporal consistency measures.}
    \label{tab:sota}
    \vspace{-0.6em}
\end{table*}

\boldparagraph{Baselines.}
We compare our method to three baselines. 

\noindent - \textit{MVDiffusion}~\cite{mvdiffusion} is a \textbf{concurrent} advancement in multi-view generative models, conditioned on depths and camera poses.
The architecture is based on LDMs~\cite{ldm} and ControlNet~\cite{controlnet}, together with the proposed correspondence-aware attention (CAA) module designed to enhance multi-view consistency.
We train the model on the HoliCity dataset by following the proposed two-stage process, \ie, training the single-view depth-conditioned diffusion model backbone first and fixing it, followed by training the ControlNet part with the CAA module for multiple views.
We use their default hyper-parameters during training but use a uniform text prompt, ``\texttt{a street with buildings}''.

\noindent - \textit{Sat2Vid}~\cite{sat2vid} was proposed to generate street panorama videos from the geometry inferred from a satellite image.
To ensure consistency across the resulting frames, it creates an explicit 3D point cloud representation for the scene and maintains its correspondence with every frame pixel.
The texture information is generated directly in 3D space by a GAN method in a 3D sparse setting.
In our experiment, we exclude their satellite stage but mainly train their generative parts on the HoliCity dataset, using the lifted \textbf{pinhole}-view point cloud and established 3D-2D correspondence map.

\noindent - \textit{InfiniCity}~\cite{infinicity} is a recent advancement for the city-scale generation based on GANcraft~\cite{gancraft}.
Since their code is not publicly available but the model is trained with the HoliCity~\cite{holicity} dataset, we directly take the FID and KID scores from their paper for reference only.

\subsection{Quantitative comparison}

The quantitative evaluation results can be found in \cref{tab:sota}.
Our method outperforms the baselines in most cases except FID and KID where we are second only to MVDiffusion~\cite{mvdiffusion}. 
Meanwhile, we hold better FVD and KVD scores, which are more comprehensive metrics for the overall video quality comparison.
This is very well aligned with the user study, where the videos generated by our method are more recognized and liked by the users.
Since MVDiffusion~\cite{mvdiffusion} is a 2D diffusion model built on top of powerful LDMs and ControlNet with billions of parameters and using a pre-trained model, the generated (single) images are therefore reasonably photorealistic.
This is reflected in their FID and KID scores that emphasize more on the reality of isolated individual frames.
\rule{0pt}{16pt}However, the video realism of image sequence not only depends on the quality of individual frames but also on the temporal consistency between them. 
Despite the proposed CAA module helping to maintain temporal consistency, it may not generate frames as consistent as rendering from a 3D representation, resulting in inferior FVD and KVD scores than our method.
This indicates that our generated frames hold very well consistency, which can be also verified in the cross-frame consistency measures PSNR, SSIM, and LPIPS, where we outperform all the baselines.
It is noteworthy that Sat2Vid~\cite{sat2vid}'s 3D-2D point-to-pixel correspondence is strictly maintained, the inferior score in consistency measures may be because their 2D upsampling module breaks the original consistency.

\subsection{Qualitative comparison}
\label{sec:qualitative}

\cref{fig:sota} includes the visualization results for six samples generated by our method and two baselines.
For MVDiffusion~\cite{mvdiffusion} and our method, we generate 48 frames for every scene with the interpolated poses between the 8 poses given by the dataset. 
Since the memory needed by Sat2Vid~\cite{sat2vid} inflates extremely with a lot of frame generation, we keep its frame number to 8 and decrease its video FPS for synchronized visualization with other 48-frame videos.
Both Sat2Vid and our method show good consistency between frames since the texture features are tiled on the scene geometry. 
Instead, while MVDiffusion connects frame texture features via geometric correspondences and limits the similarity, it may still struggle to guarantee complete consistency between frames.
This challenge could become particularly severe for 2D methods when dealing with a smaller overlap ratio in the training data.
For instance, the average foreground overlap between two neighboring ``key'' frames is $\sim$42\% in HoliCity~\cite{holicity}, as opposed to $\sim$65\% in ScanNet~\cite{scannet} used in MVDiffusion, which could be the primary reason that it presents an inferior consistency compared with the results reported in its original paper.
When considering photorealism, our method and MVDiffusion generate more plausible textures for the building facades than Sat2Vid.
MVDiffusion can output more diversified building textures than ours.
We infer this is because the LDMs~\cite{ldm} embedded in MVDiffusion have a higher ability to memorize and generate texture patterns with its large number of pre-trained parameters, compared with the diffusion model used by our method.
In short, ours achieves a commendable equilibrium between the photorealism and consistency of synthesized results, showcasing a balance superior to Sat2Vid and MVDiffusion.

\begin{table}[b]
    \centering
    \vspace{-1em}
    \begin{tabular}{l|cc|c}
        \toprule
        \textbf{Variant / Metric} & FID $\downarrow$ & KID$_{\times\text{100}}$ $\downarrow$ & Dep. RMSE \\
        \midrule
        w/o pt-rsmp & 131.38 & 12.66$^{\pm\text{0.12}}$ & - \\
        w/o pt-aggr & 85.58 & 7.79$^{\pm\text{0.08}}$ & 3.22 \\
        w/o dep-sup & 80.40 & 7.22$^{\pm\text{0.08}}$ & 3.44 \\
        \midrule
        \textbf{Ours} & \textbf{71.98} & \textbf{5.91}$^{\pm\text{0.06}}$ & \textbf{3.07} \\
        \bottomrule
    \end{tabular}
    \caption{\textbf{Ablative evaluation of our method.} The result shows the effectiveness of each component.}
    \label{tab:ablation}
\end{table}

\subsection{Ablation study}
\label{sec:ablation}

\begin{figure}[!hb]
    \centering
    \vspace{-0.75em}
    \footnotesize
    \newcommand{\sz}{0.12\textwidth} %

    \setlength{\tabcolsep}{1pt}
    \begin{tabular}{C{\sz}C{\sz}C{\sz}C{\sz}}
    w/o pt-rsmp & w/o pt-aggr & w/o dep-sup & Ours \\
    
    \includegraphics[width=\sz]{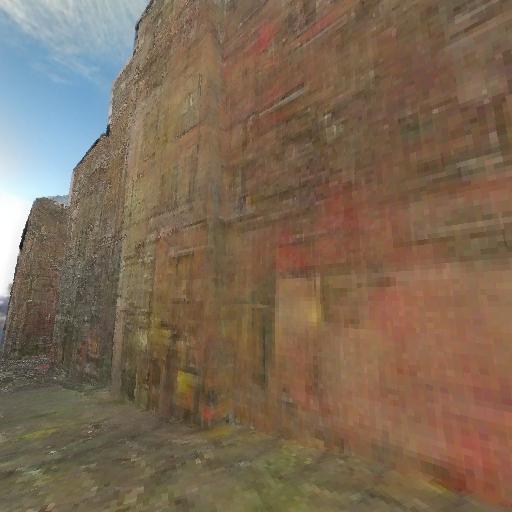} &
    \includegraphics[width=\sz]{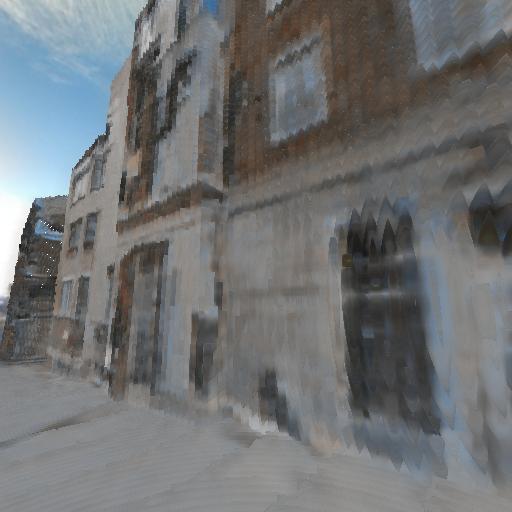} &
    \includegraphics[width=\sz]{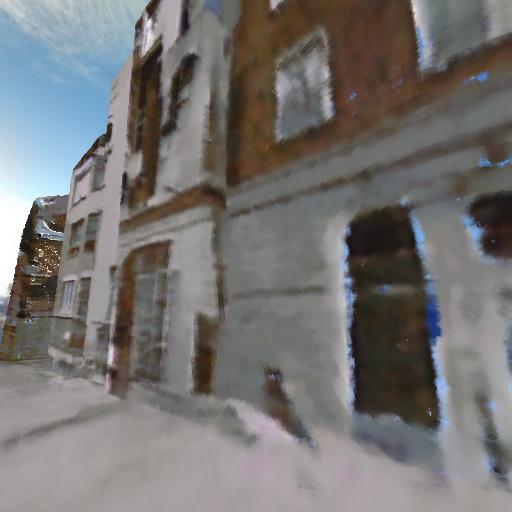} &
    \includegraphics[width=\sz]{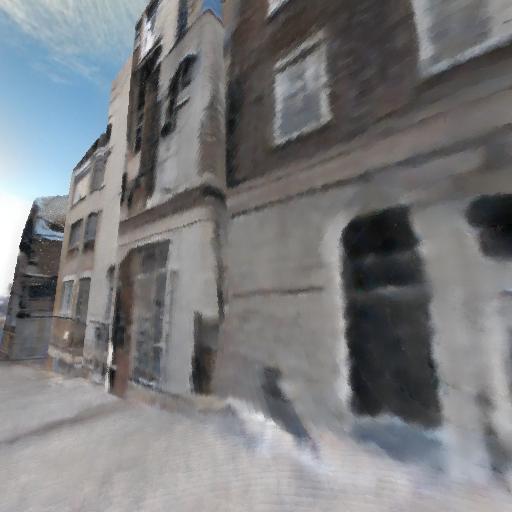} \\

    \includegraphics[width=\sz]{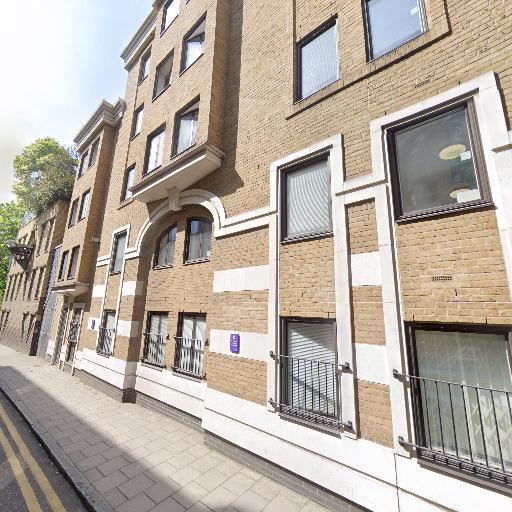} &
    \includegraphics[width=\sz]{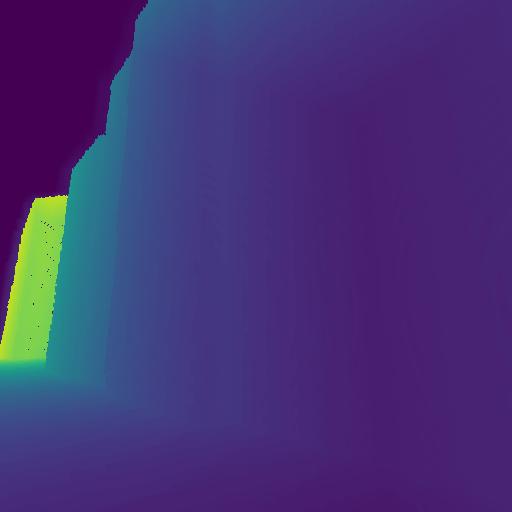} &
    \includegraphics[width=\sz]{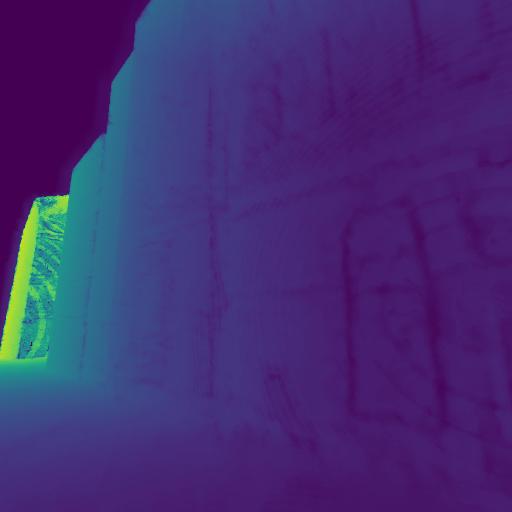} &
    \includegraphics[width=\sz]{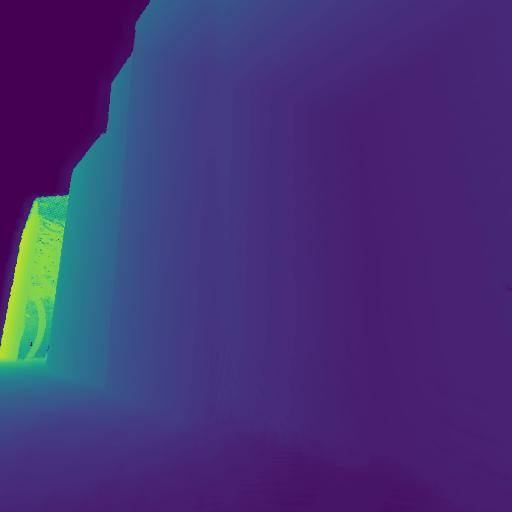} \\

    \midrule

    \includegraphics[width=\sz]{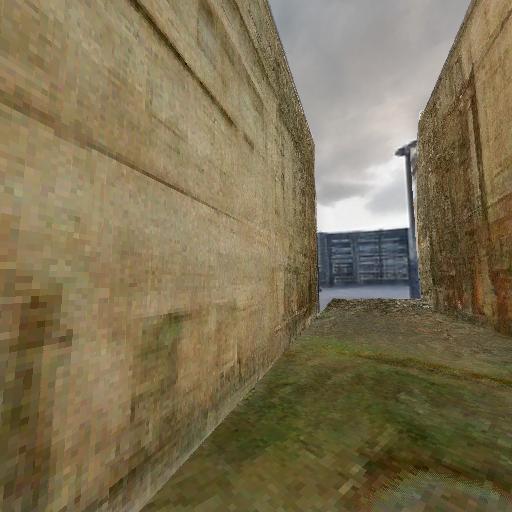} &
    \includegraphics[width=\sz]{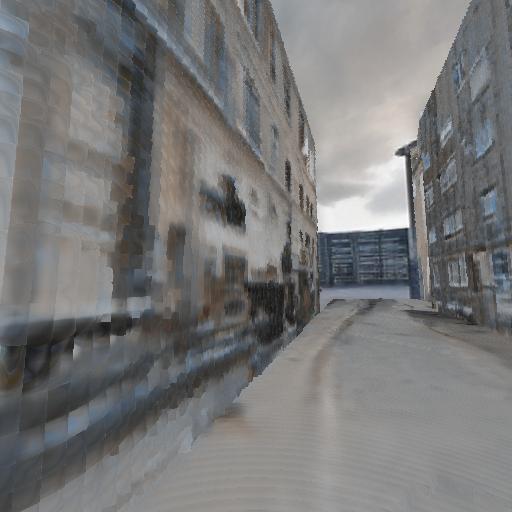} &
    \includegraphics[width=\sz]{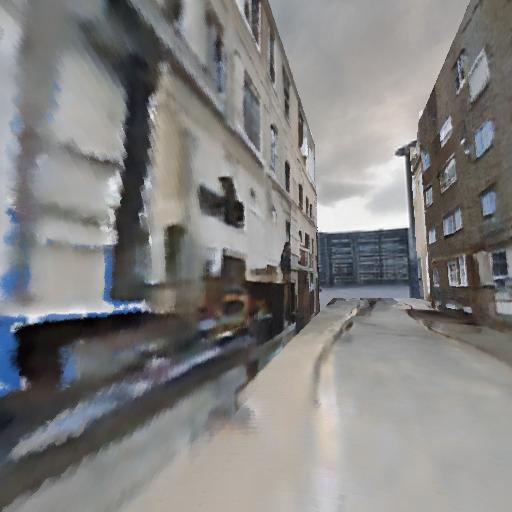} &
    \includegraphics[width=\sz]{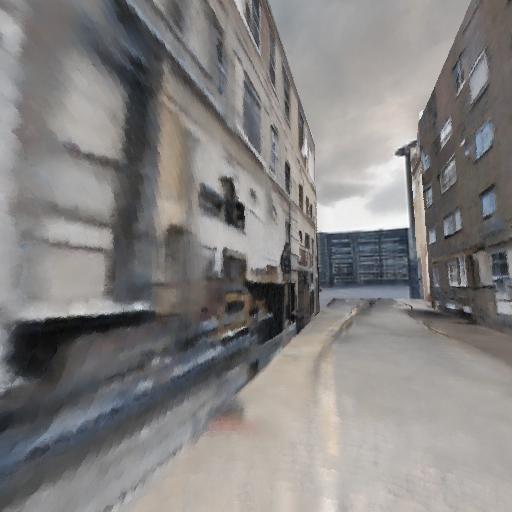} \\

    \includegraphics[width=\sz]{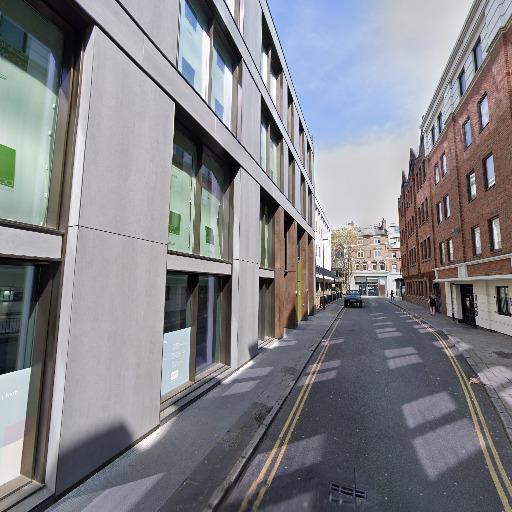} &
    \includegraphics[width=\sz]{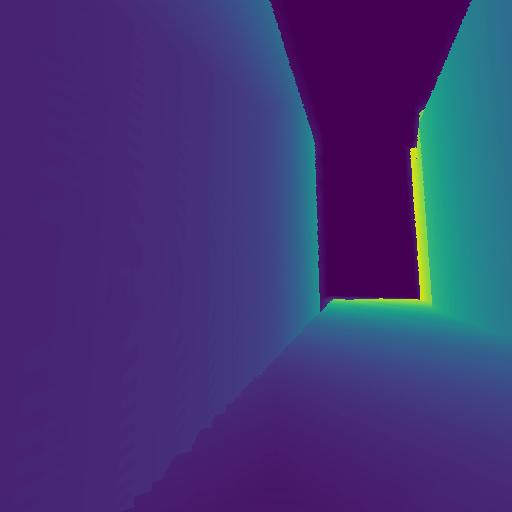} &
    \includegraphics[width=\sz]{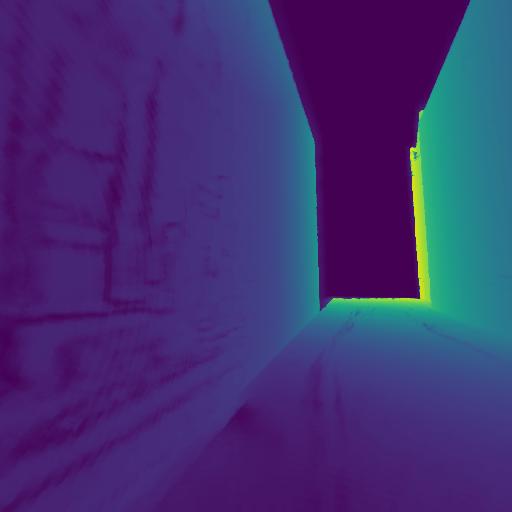} &
    \includegraphics[width=\sz]{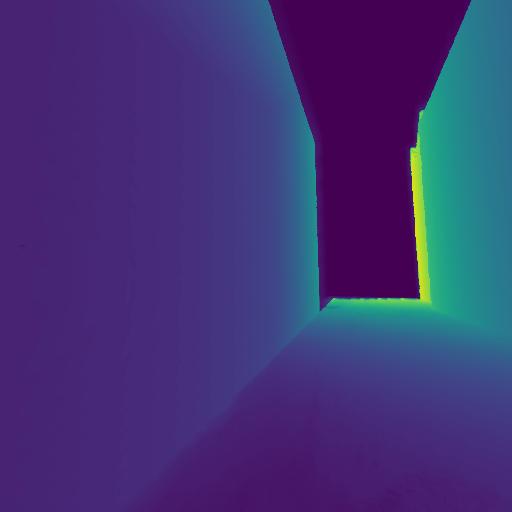} \\

    \midrule

    \includegraphics[width=\sz]{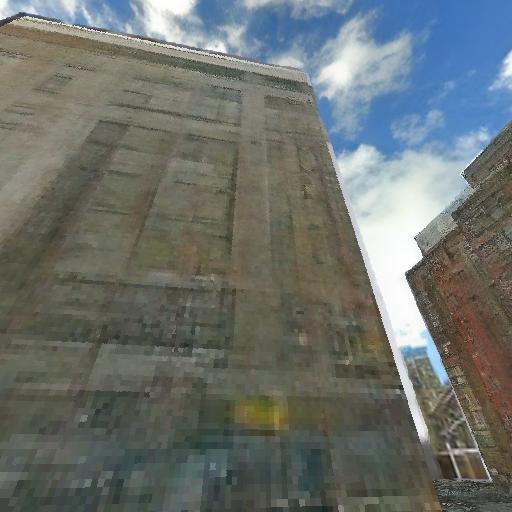} &
    \includegraphics[width=\sz]{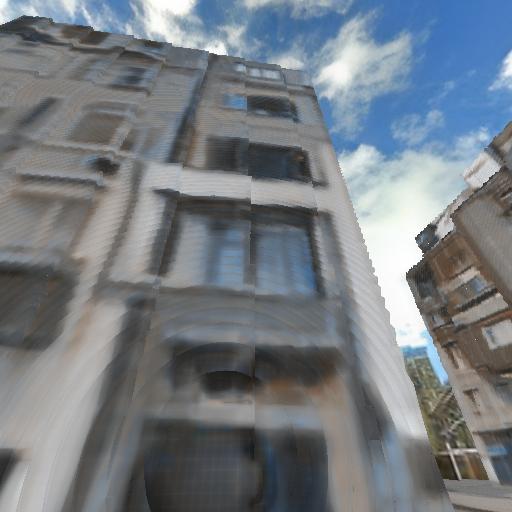} &
    \includegraphics[width=\sz]{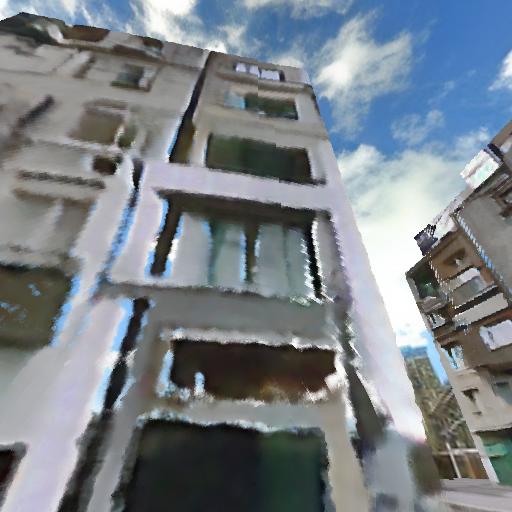} &
    \includegraphics[width=\sz]{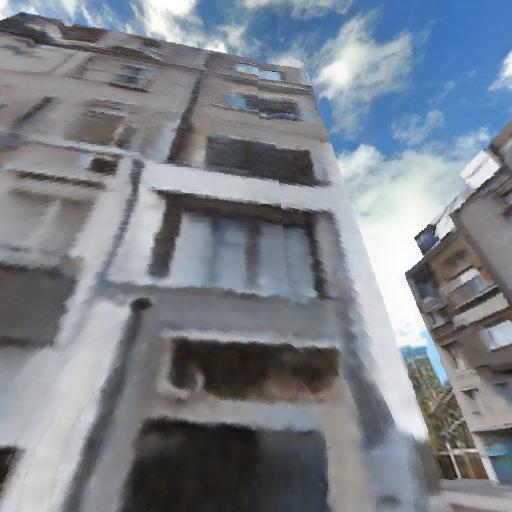} \\
    
    \includegraphics[width=\sz]{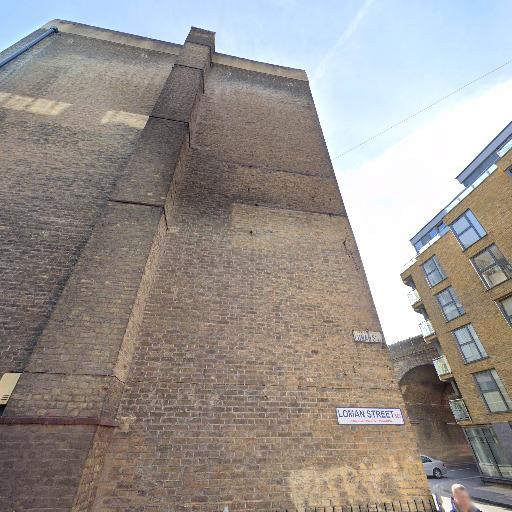} &
    \includegraphics[width=\sz]{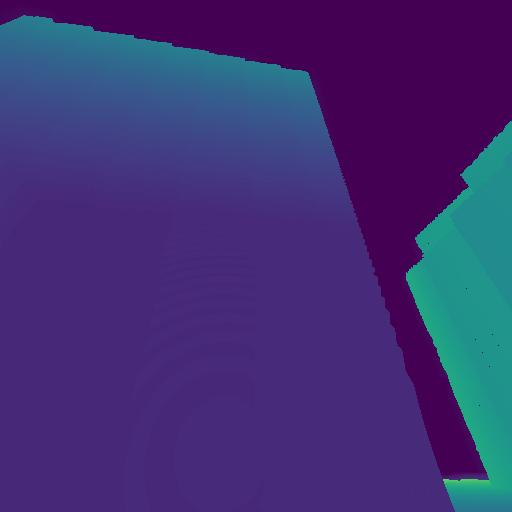} &
    \includegraphics[width=\sz]{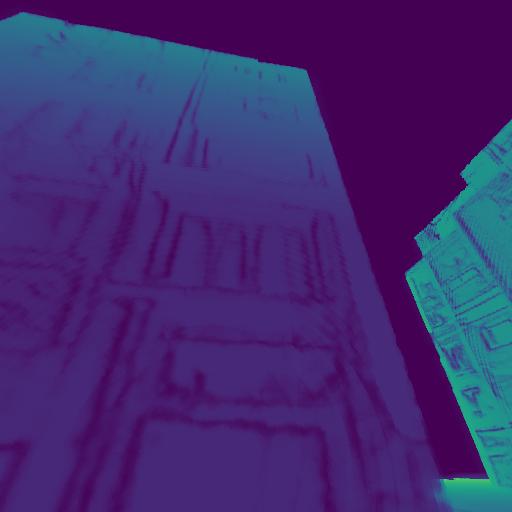} &
    \includegraphics[width=\sz]{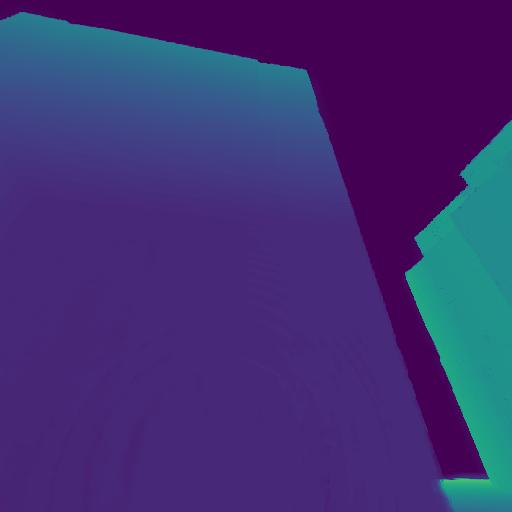} \\
    
    \end{tabular}
    
    \caption{\textbf{Qualitative ablation study.} We present exemplary qualitative results for various ablations of our method. The rendered images visibly contain more details and the depths are recovered better with our full method. \textbf{The second line of each example shows the depth in pseudo colors, except the bottom left ones which are GT images.}}
    \label{fig:ablation}
\end{figure}

We further conduct ablative experiments to validate the effectiveness of the three components utilized in our method.
\begin{itemize}
    \item \textbf{w/o pt-rsmp}: Do not use point resampling for balancing the point density of point clouds in the generation phase as mentioned in \cref{sec:impl}, use the pinhole views back-projection; The 3D diffusion model is directly trained on this structure and no neural rendering in this variant as we can directly take the colors from the denoised point cloud;
    \item \textbf{w/o pt-aggr}: Do not use point aggregation in the rendering phase, use voxel-based interpolation from the output of the feature extractor;
    \item \textbf{w/o dep-sup}: Do depth supervision in training the neural renderer, loss only calculated from color difference.
\end{itemize}
We remove each operation individually from our full model training procedure and present the quantitative evaluation results in \cref{tab:ablation} and the visualization results in \cref{fig:ablation}.
Since the inter-frame consistency performance of each method is very similar, we only compare the quality measures FID and KID, and further calculate the RMSE value of the depth map (except w/o pt-rsmp because it always gets groud-truth depth).
It can be observed that removing any of these operations results in a decline in image quality, as evidenced by both metrics and visualization.
In particular, the omission of the point resampling scheme leads to a dramatic decrease in resulting image quality. 
This underscores the critical importance of maintaining an even density in the point cloud for the diffusion model's denoising process to effectively generate meaningful textures.
Without the use of point aggregation, the features are organized in a slightly lower resolution and thus the representations are more coarse, leading to a decline in the rendered image quality and increased blurriness.
As shown in \cref{fig:ablation}, the results without depth supervision consistently exhibit holes in the rendered frames, particularly for buildings near the camera. 
While more refined depths for patterns on the buildings, such as windows and doors, can be generated, their geometries are not well-connected.
This results in the ray hitting directly into the sky in these open areas and generating holes in the images.

\begin{figure}[b]    
    \centering
    \vspace{-0.75em}
    \footnotesize
    
    \setlength{\tabcolsep}{2pt}
    \begin{tabular}{C{0.147\textwidth}|L{0.30\textwidth}l}
     Satellite & \hspace{1em}MVDiffusion~\cite{mvdiffusion}\hspace{4.5em}Ours & \\

    \embedvideo*{\includegraphics[width=0.147\textwidth]{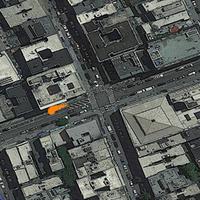}}{media/omnicity/sat_0.mp4} \includegraphics[width=0.147\textwidth]{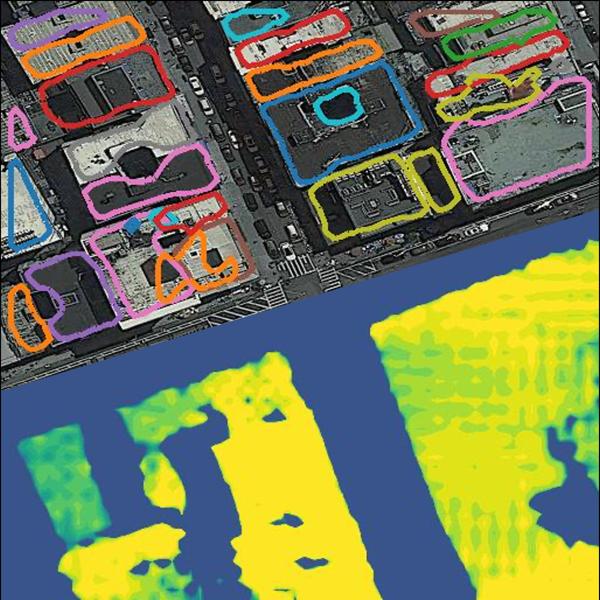} &
    \embedvideo*{\includegraphics[width=0.3\textwidth]{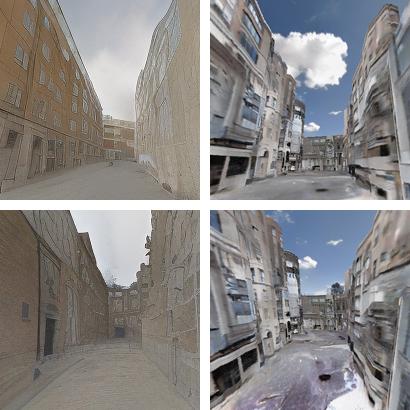}}{media/omnicity/omnicity_0_b800k.mp4} &
    \hspace{0em}\rotatebox[origin=c]{90}{\hspace{0em}Bird-view\hspace{5em}Ground-view\hspace{0em}} \\
    
    \embedvideo*{\includegraphics[width=0.145\textwidth]{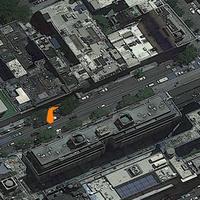}}{media/omnicity/sat_3.mp4} \includegraphics[width=0.147\textwidth]{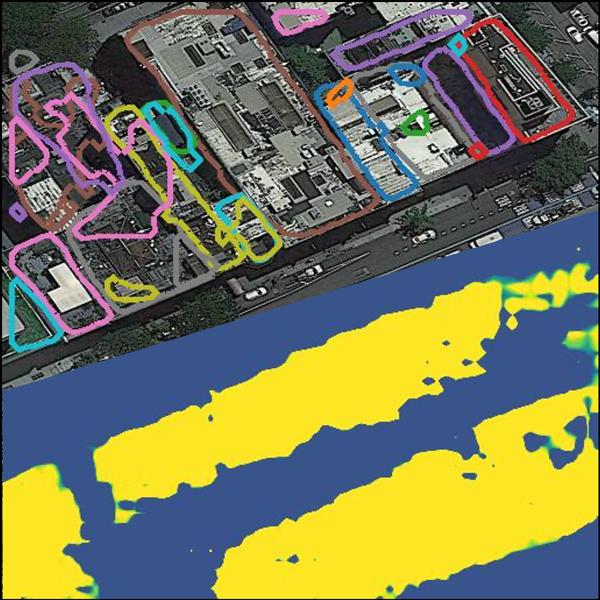} &
    \embedvideo*{\includegraphics[width=0.3\textwidth]{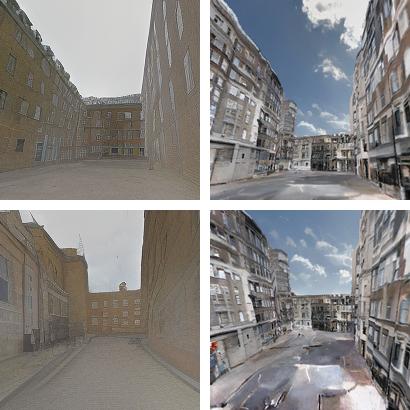}}{media/omnicity/omnicity_3_b800k.mp4} & 
    \hspace{0em}\rotatebox[origin=c]{90}{\hspace{0em}Bird-view\hspace{5em}Ground-view\hspace{0em}} \\
 
    \end{tabular}
    \caption{\textbf{Street-view videos generated on the OmniCity~\cite{omnicity} dataset.} Ours shows better generalization ability, whereas MVDiffusion~\cite{mvdiffusion} generates bird-view videos as from ground views.\VIDEO}
    \label{fig:omnicity}
\end{figure}

\subsection{Generalization}
\label{sec:generalization}

In order to demonstrate the usability of our pipeline on the satellite-to-ground cross-view generation task, and highlight its generalization capabilities, we conducted additional experiments on the OmniCity~\cite{omnicity} dataset.
Specifically, we utilize the pre-trained monocular depth estimation network DRON~\cite{dron} and the instance segmentation model Mask R-CNN~\cite{maskrcnn} as provided by~\cite{omnicity}.
Exemplary prediction results of the two networks are shown in the left column of \cref{fig:omnicity}.
The scene geometry is inferred from the combined results of the two models: the top-view height map and building footprints.
For each scene, we create ground-view and bird-view navigation trajectories along the roads, containing around 500 frames.
We then generate videos along the trajectories using the model trained on the HoliCity dataset.
We compare the generation qualitatively on selected scenes with MVDiffusion~\cite{mvdiffusion} and show the videos in the right columns of ~\cref{fig:omnicity}.
Our method consistently produces frames with superior consistency across different positions and view angles.
In contrast, MVDiffusion generates images with varied textures even for the same position under different view angles.
Moreover, in the bird-view generation, the results produced by MVDiffusion are still biased towards street-view images.
This is probably because the depth image used as input to the model follows the bird-view characteristics, which is out-of-distribution with the street-view training set used to train MVDiffusion, thus failing to show good generalization ability.
Since our model operates directly in 3D space, the bird-view images can be still rendered even if these views are not fed to the network during the training process.

\section{Conclusion}
\label{sec:conclu}

In this paper, we explored direct scene generation from satellite imagery.
Our novel architecture brings diffusion models into 3D sparse representations and integrates them with neural rendering techniques.
This approach involves generating texture colors at the point level using the proposed 3D diffusion model, subsequently converting it into a scene representation without test-time optimization.
The resulting representation allows the rendering of arbitrary views with excellence in both single-frame quality and inter-frame consistency.
Experimental results demonstrate the superiority of our model in generating photorealistic street-view videos.
To the best of our knowledge, we presented the first work that utilizes the diffusion model in a 3D sparse setting for cross-view urban scene generation from satellite imagery.
We regard our work as a step forward in overcoming the challenges associated with urban scene generation, offering a promising avenue for future developments in realistic and adaptable visual content synthesis.

\vspace{1em}

\boldparagraph{Acknowledgement}.
Zuoyue Li was supported by the Swiss Data Science Center (SDSC) fellowship program. Zhaopeng Cui was affiliated with the State Key Lab of CAD \& CG at Zhejiang University. Martin R. Oswald was supported by a FIFA research grant.

\clearpage

{
    \small
    \bibliographystyle{ieeenat_fullname}
    \bibliography{main}
}

\clearpage

\ExplSyntaxOn
\NewDocumentCommand\embedvideosupp{smm}{
  \group_begin:
  \leavevmode
  \tl_if_exist:cTF{file_\file_mdfive_hash:n{#3}}{
    \tl_set_eq:Nc\video{file_\file_mdfive_hash:n{#3}}
  }{
    \IfFileExists{#3}{}{\GenericError{}{File~`#3'~not~found}{}{}}
    \pbs_pdfobj:nnn{}{fstream}{{}{#3}}
    \pbs_pdfobj:nnn{}{dict}{
      /Type/Filespec/F~(#3)/UF~(#3)
      /EF~<</F~\pbs_pdflastobj:>>
    }
    \tl_set:Nx\video{\pbs_pdflastobj:}
    \tl_gset_eq:cN{file_\file_mdfive_hash:n{#3}}\video
  }
  \pbs_pdfobj:nnn{}{dict}{
    /Type/RichMediaInstance/Subtype/Video
    /Asset~\video
    /Params~<</FlashVars (
    )>>
  }
  \pbs_pdfobj:nnn{}{dict}{
    /Type/RichMediaConfiguration/Subtype/Video
    /Instances~[\pbs_pdflastobj:]
  }
  \pbs_pdfobj:nnn{}{dict}{
    /Type/RichMediaContent
    /Assets~<<
      /Names~[(#3)~\video]
    >>
    /Configurations~[\pbs_pdflastobj:]
  }
  \tl_set:Nx\rmcontent{\pbs_pdflastobj:}
  \pbs_pdfobj:nnn{}{dict}{
    /Activation~<<
      /Condition/\IfBooleanTF{#1}{PV}{XA}
      /Presentation~<</Style/Embedded>>
    >>
    /Deactivation~<</Condition/PI>>
  }
  \hbox_set:Nn\l_tmpa_box{#2}
  \tl_set:Nx\l_box_wd_tl{\dim_use:N\box_wd:N\l_tmpa_box}
  \tl_set:Nx\l_box_ht_tl{\dim_use:N\box_ht:N\l_tmpa_box}
  \tl_set:Nx\l_box_dp_tl{\dim_use:N\box_dp:N\l_tmpa_box}
  \pbs_pdfxform:nnnnn{1}{1}{}{}{\l_tmpa_box}
  \pbs_pdfannot:nnnn{\l_box_wd_tl}{\l_box_ht_tl}{\l_box_dp_tl}{
    /Subtype/RichMedia
    /BS~<</W~0/S/S>>
    /Contents~(embedded~video~file:#3)
    /NM~(rma:#3)
    /AP~<</N~\pbs_pdflastxform:>>
    /RichMediaSettings~\pbs_pdflastobj:
    /RichMediaContent~\rmcontent
  }
  \phantom{#2}
  \group_end:
}
\ExplSyntaxOff

\twocolumn[{%
  \renewcommand\twocolumn[1][]{#1}%
  \maketitlesupplementary
  \centering
\footnotesize
\newcommand{\sz}{0.32\textwidth}
\setlength{\tabcolsep}{4pt}
\begin{tabular}{C{\sz}|C{\sz}|C{\sz}}
\embedvideosupp*{\includegraphics[width=\sz]{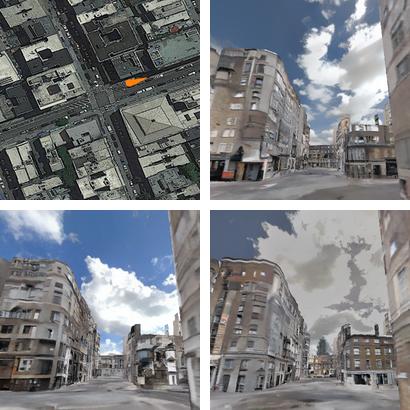}}{media/style_videos/style_0.mp4} & 
\embedvideosupp*{\includegraphics[width=\sz]{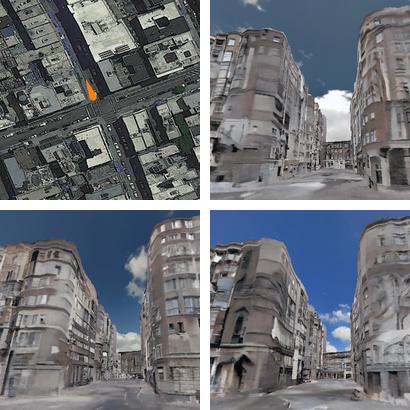}}{media/style_videos/style_1.mp4} & 
\embedvideosupp*{\includegraphics[width=\sz]{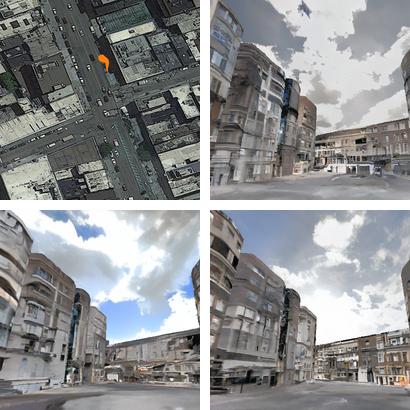}}{media/style_videos/style_2.mp4} \\
Scene 1 & Scene 2 & Scene 3
\end{tabular}
\captionof{figure}{\textbf{Denoising procedure visualization with three different diversities}. We align the denoising processes of background and foreground for better visual effects. For each scene, we also provide its satellite top view in the upper-left corner with current locations.\VIDEO\VIDEOREMINDER}
\label{fig:style}
\vspace{2em}

}]

In this supplementary material, we provide further details (\cref{sec:supp_details}), present additional experiment results (\cref{sec:supp_exp}), and discuss the limitations of our method along with potential future directions (\cref{sec:supp_lim}).

\section{Additional details}
\label{sec:supp_details}

\boldparagraph{OmniCity~\cite{omnicity} video poses.}
All the videos in \cref{fig:teaser}, \cref{fig:omnicity}, and \cref{fig:style} use trajectories along the right side of the road. 
The ground-view poses are at a height level of 2m, 15$^{\circ}$ pitch facing the sky and zero roll, while bird-view poses are at a height level of 10m, 15$^{\circ}$ pitch facing the road and zero roll.

\boldparagraph{Data augmentation.}
Since the number of scenes ($\sim$5k) of the HoliCity~\cite{holicity} dataset is not sufficient enough, we add the following data augmentations during training. 
We randomly flip the whole scene along the two horizontal axes, while we also randomly rotate the scene around the vertical axis.
In addition, we perform Gamma correction on the point cloud RGB ground truth, with a random $\gamma$ value between 0.8 and 1.25.

\boldparagraph{Inference timing.}
The generation of the 3D sparse diffusion models takes approximately 10 minutes for the whole denoising process of a single scene. 
For the neural rendering phase, the inference takes around 1.6s per frame with a resolution of 512$\times$512.

\boldparagraph{Other related baselines.}
We did not include experiments with Vid2Vid~\cite{vid2vid}, WC-Vid2Vid~\cite{wcvid2vid} and Sat2Density~\cite{sat2density} due to the following reasons.
Vid2Vid~\cite{vid2vid} has already exhibited deficiencies in temporal consistency, as observed in Sat2Vid~\cite{sat2vid}.
WC-Vid2Vid~\cite{wcvid2vid}, on the other hand, necessitates semantic video input, which exceeds the scope of our problem setting. 
Furthermore, Sat2Density~\cite{sat2density} requires for precise satellite-ground image correspondence, which is unavailable in the HoliCity~\cite{holicity} dataset.

\section{Additional experiment results}
\label{sec:supp_exp}

\begin{figure*}[!ht]
    \centering
    \footnotesize
    \begin{tabular}{C{0.0000001\textwidth}C{0.97\textwidth}}
        \rotatebox{90}{GT} & \embedvideosupp*{\includegraphics[width=0.97\textwidth]{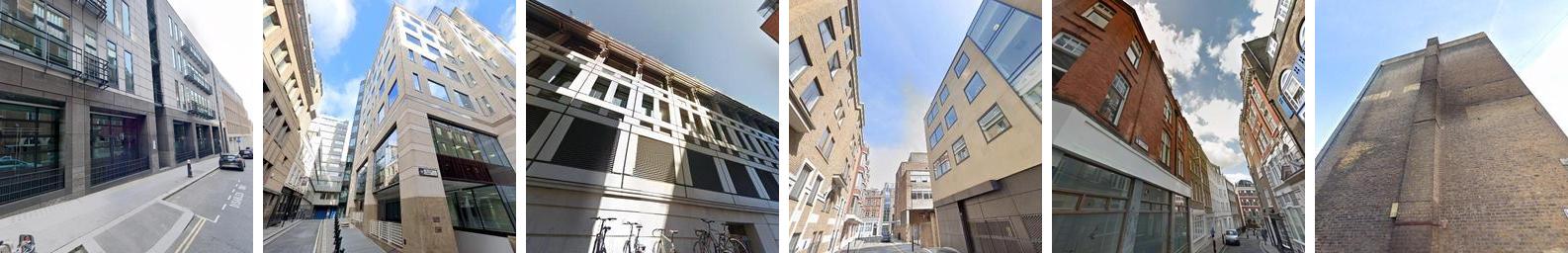}}{media/supp_baseline/supp_GT_6res.mp4} \\
        \rotatebox{90}{Sat2Vid~\cite{sat2vid}} & \embedvideosupp*{\includegraphics[width=0.97\textwidth]{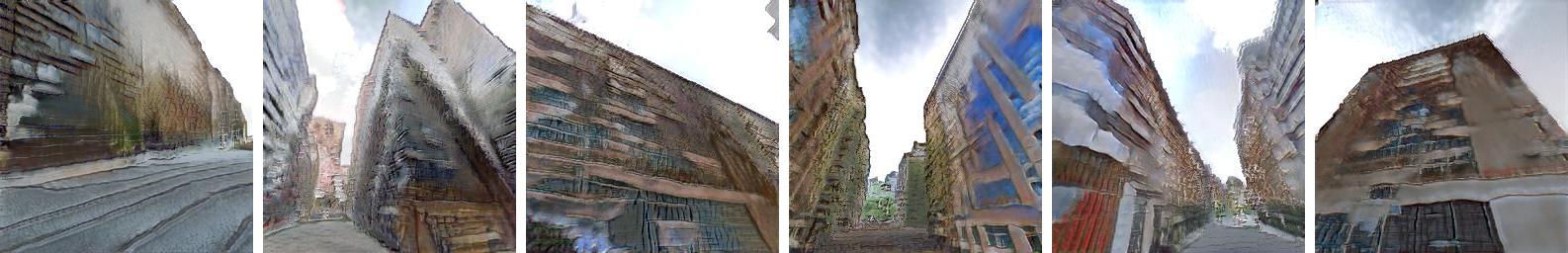}}{media/supp_baseline/supp_Sat2Vid_6res.mp4} \\
        \rotatebox{90}{MVDiffusion~\cite{mvdiffusion}} & \embedvideosupp*{\includegraphics[width=0.97\textwidth]{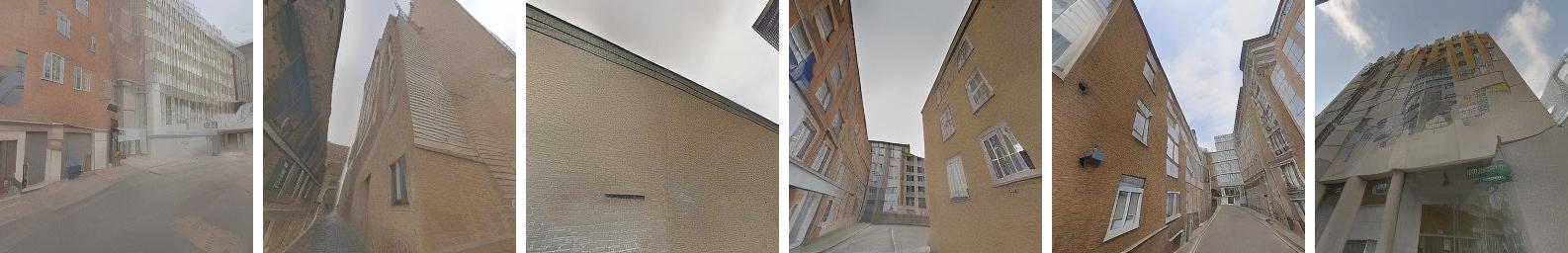}}{media/supp_baseline/supp_MVDiff_6res.mp4} \\
        \rotatebox{90}{Ours} & \embedvideosupp*{\includegraphics[width=0.97\textwidth]{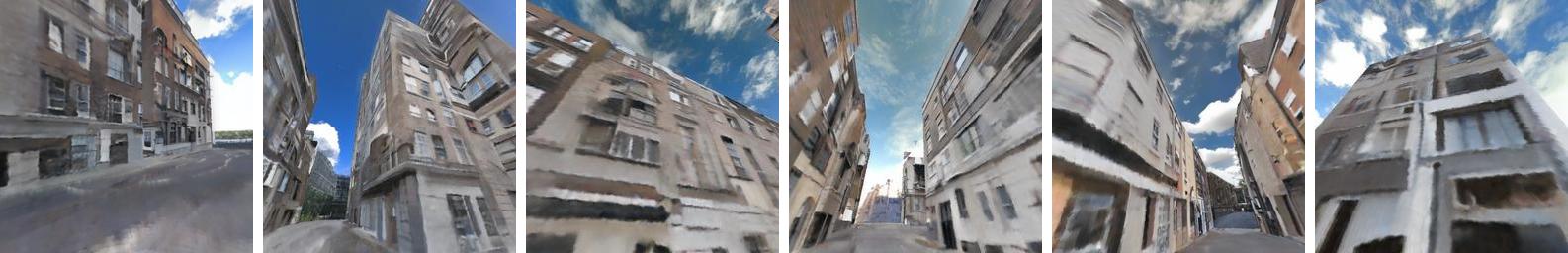}}{media/supp_baseline/supp_Ours_6res.mp4} \\
    \end{tabular}
    \caption{\textbf{Additional qualitative baseline comparison on the HoliCity~\cite{holicity} dataset.} Our method produces higher-quality videos with better temporal consistency compared to the baselines.\VIDEO}
    \label{fig:sota_supp}
\end{figure*}

\boldparagraph{Visualization of the denoising processes} on the exemplary scenes are presented in \cref{fig:style}. 
In the initial seconds of the videos, the transformation of texture from complete noise to meaningful visual patterns is evident. 
Larger noise patches are observed in the sky background, and the denoising process exhibits stability due to the utilization of LDMs~\cite{ldm}.

\boldparagraph{Multi-style generation} examples are also shown in \cref{fig:style}.
For each scene, we present three different diversities which are denoised from different noise seeds.
It is clear that our model can generate different styles of texture for the same geometry.
We noticed that no matter how the style varies, the number of floors of the generated building facade remains similar (approximately 3m per floor) as long as the building height is constant.
Also, the style of the ground floor is generally different from that of the higher floors, which is also consistent with real-world buildings.

\begin{figure}[!t]
    \centering
    \vspace{-4.5em}
    \footnotesize
    \newcommand{\sz}{0.12\textwidth} %

    \setlength{\tabcolsep}{1pt}
    \begin{tabular}{C{\sz}C{\sz}C{\sz}C{\sz}}
    w/o pt-rsmp & w/o pt-aggr & w/o dep-sup & Ours \\
    
    \includegraphics[width=\sz]{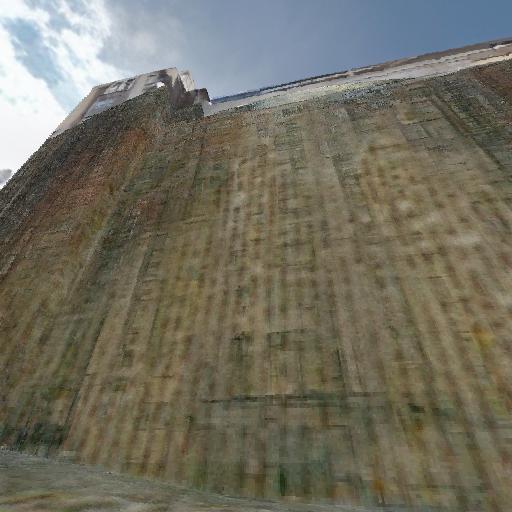} &
    \includegraphics[width=\sz]{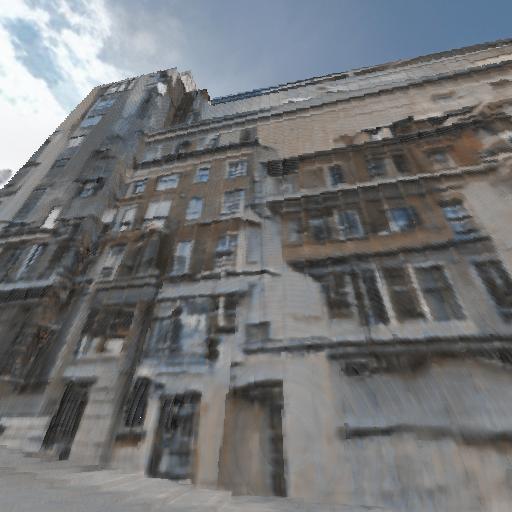} &
    \includegraphics[width=\sz]{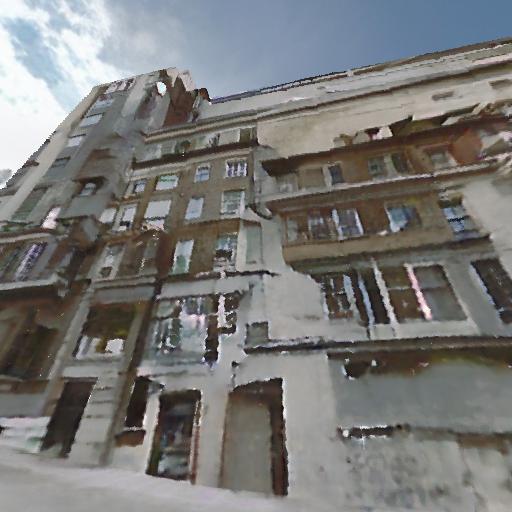} &
    \includegraphics[width=\sz]{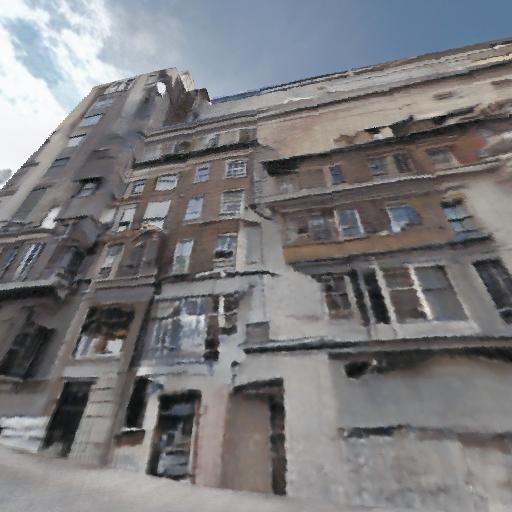} \\

    \includegraphics[width=\sz]{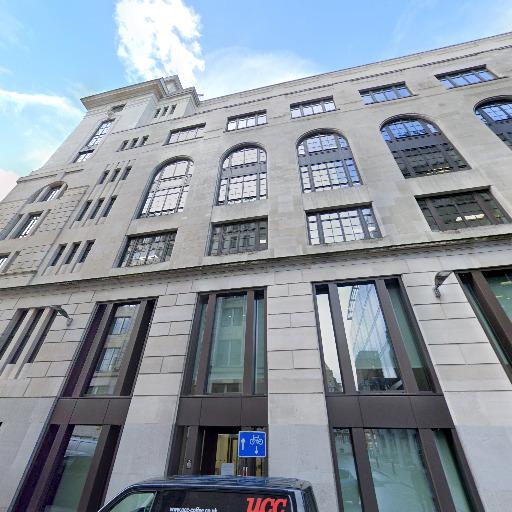} &
    \includegraphics[width=\sz]{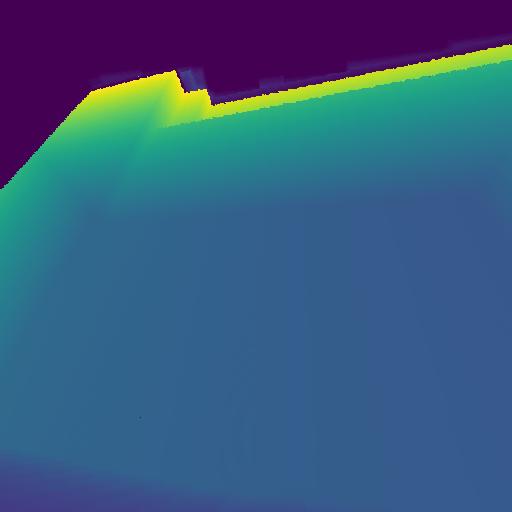} &
    \includegraphics[width=\sz]{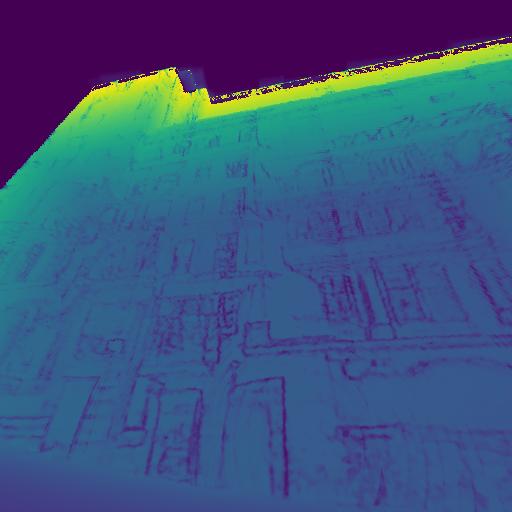} &
    \includegraphics[width=\sz]{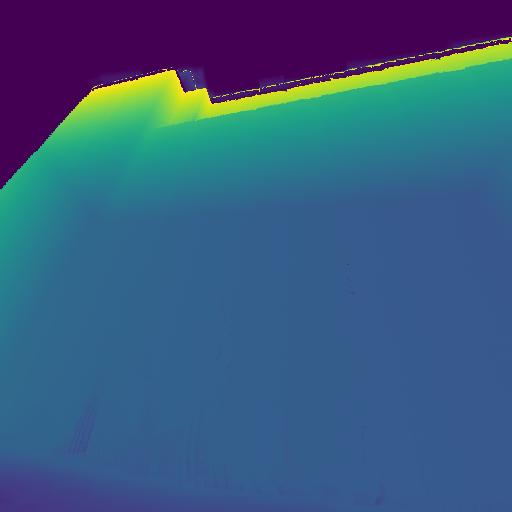} \\

    \midrule

    \includegraphics[width=\sz]{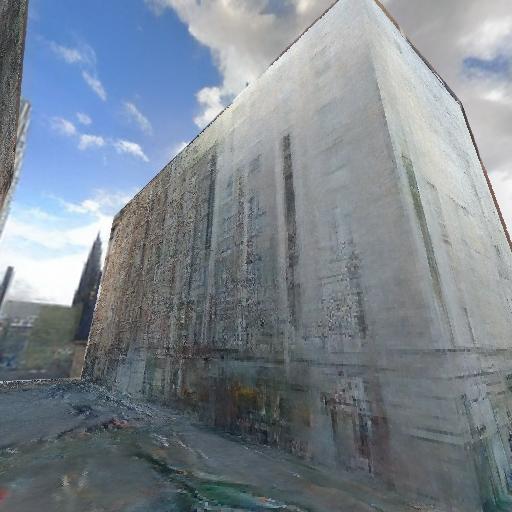} &
    \includegraphics[width=\sz]{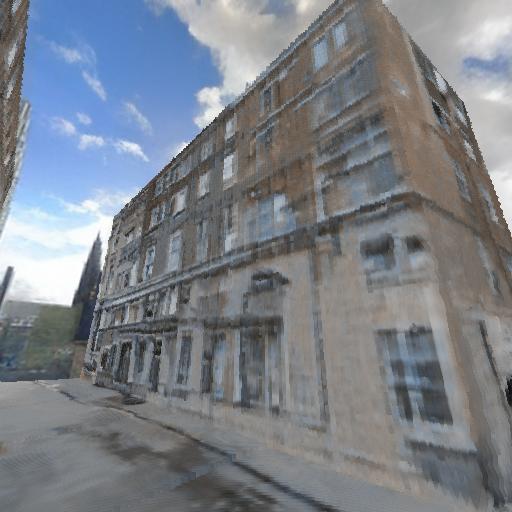} &
    \includegraphics[width=\sz]{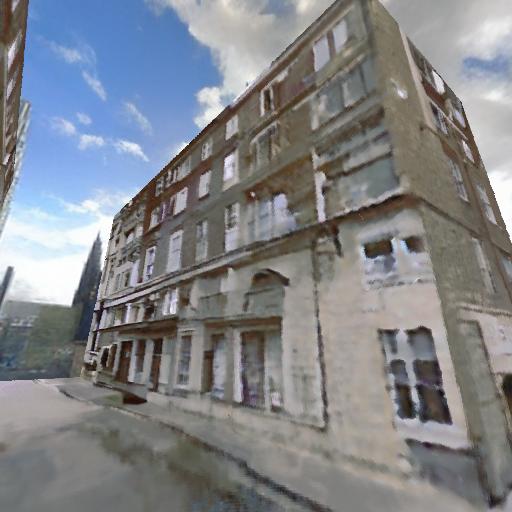} &
    \includegraphics[width=\sz]{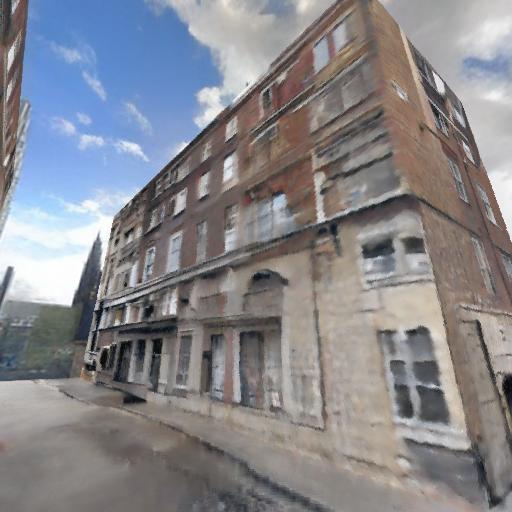} \\

    \includegraphics[width=\sz]{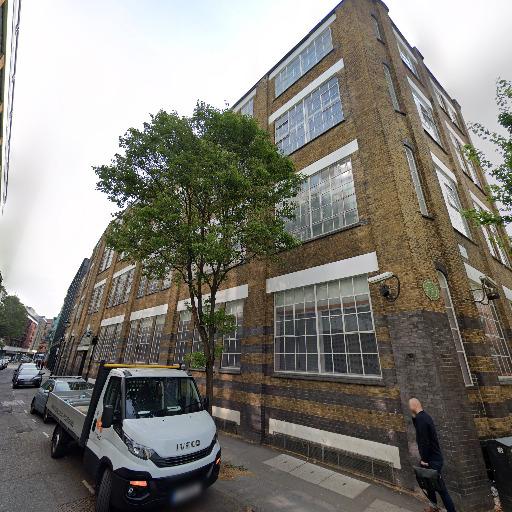} &
    \includegraphics[width=\sz]{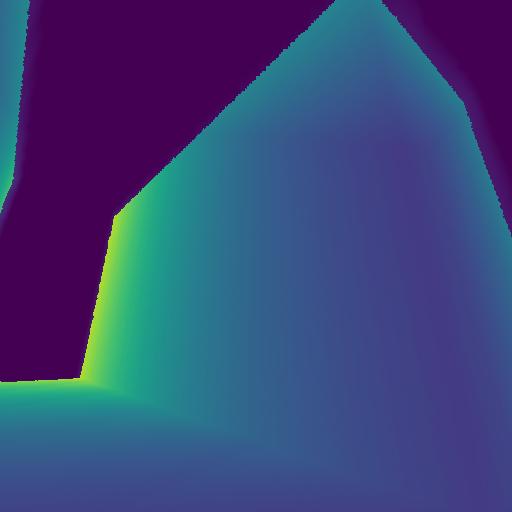} &
    \includegraphics[width=\sz]{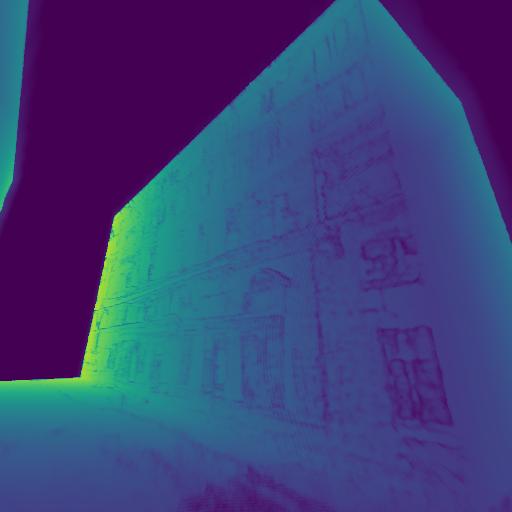} &
    \includegraphics[width=\sz]{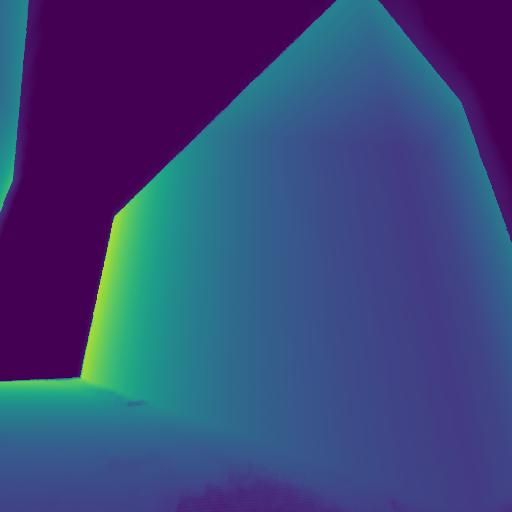} \\

    \midrule

    \includegraphics[width=\sz]{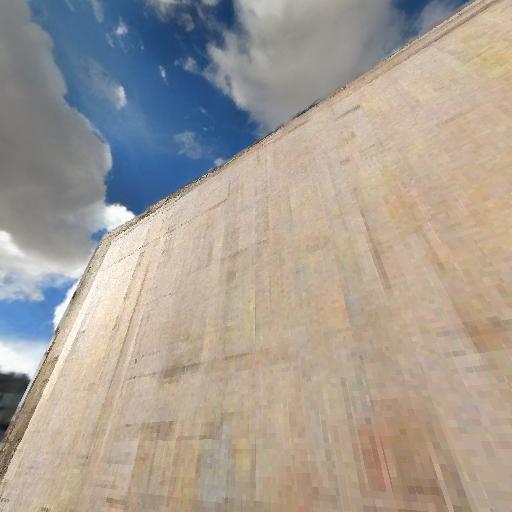} &
    \includegraphics[width=\sz]{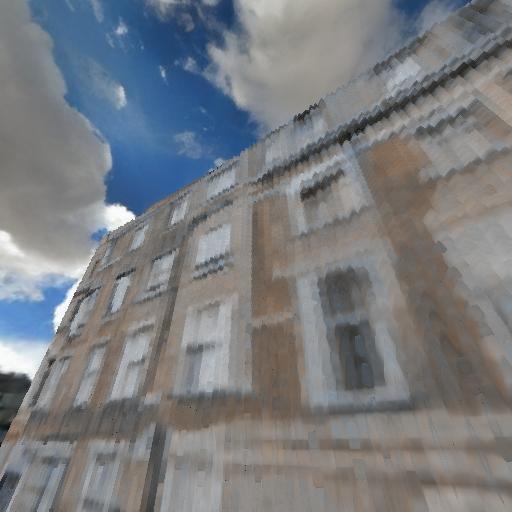} &
    \includegraphics[width=\sz]{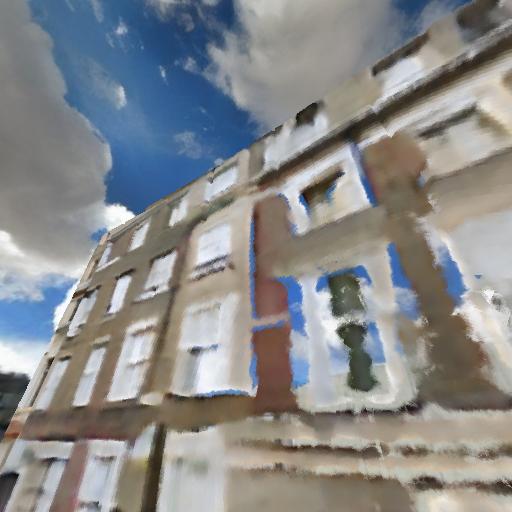} &
    \includegraphics[width=\sz]{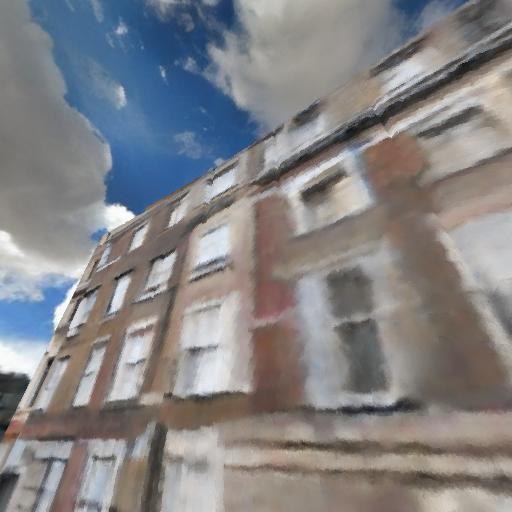} \\
    
    \includegraphics[width=\sz]{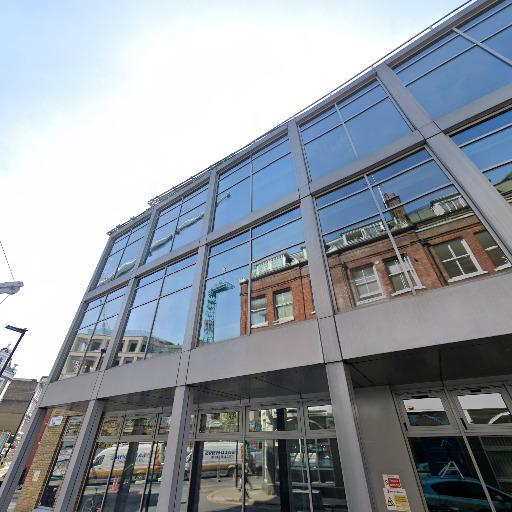} &
    \includegraphics[width=\sz]{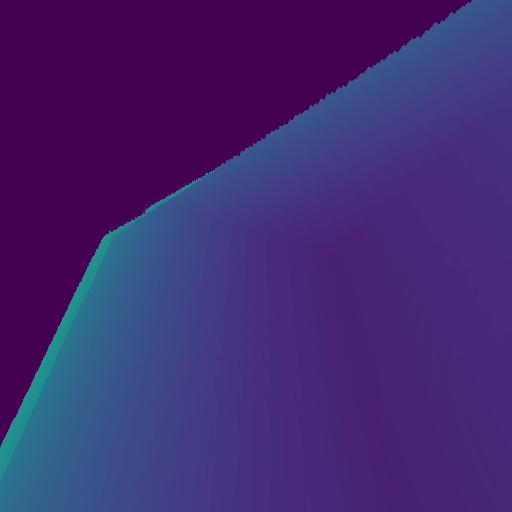} &
    \includegraphics[width=\sz]{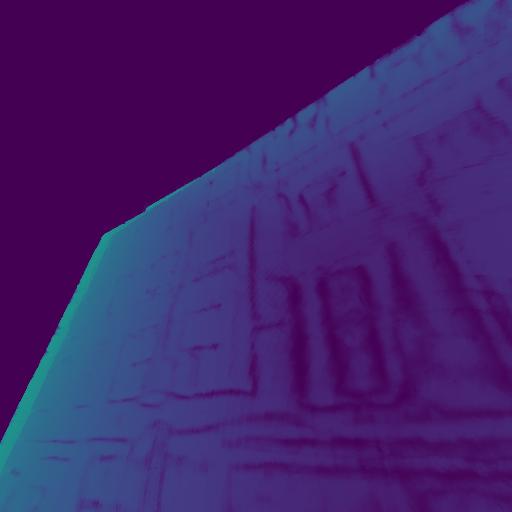} &
    \includegraphics[width=\sz]{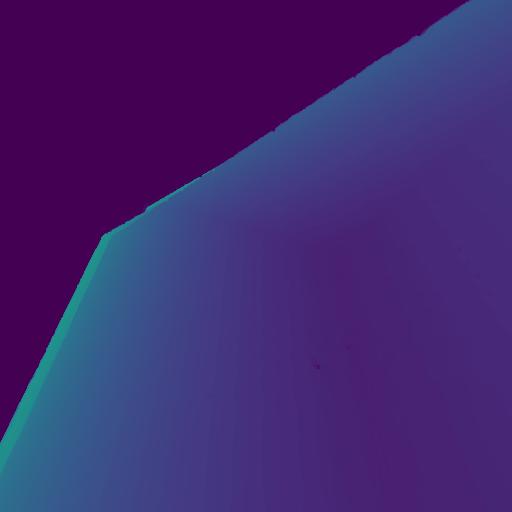} \\

    \midrule

    \includegraphics[width=\sz]{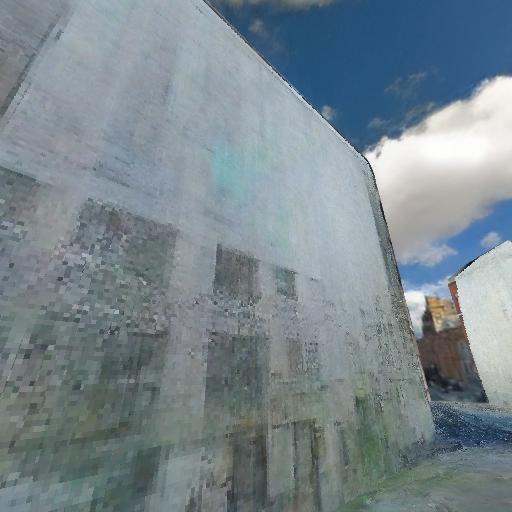} &
    \includegraphics[width=\sz]{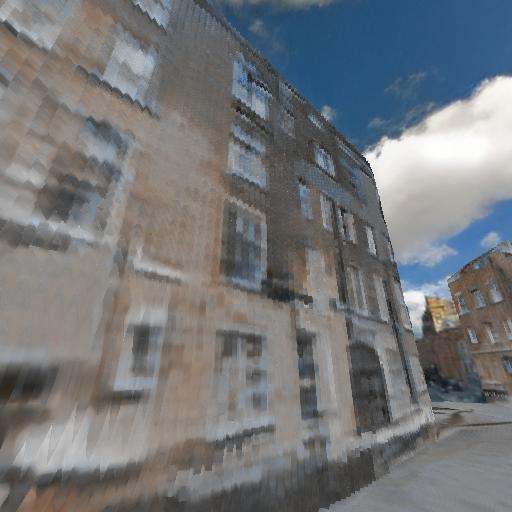} &
    \includegraphics[width=\sz]{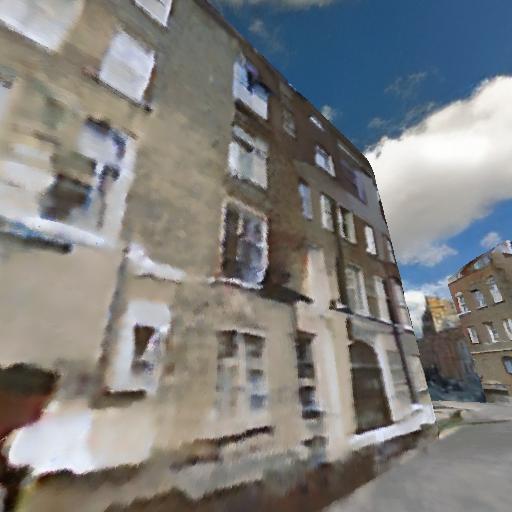} &
    \includegraphics[width=\sz]{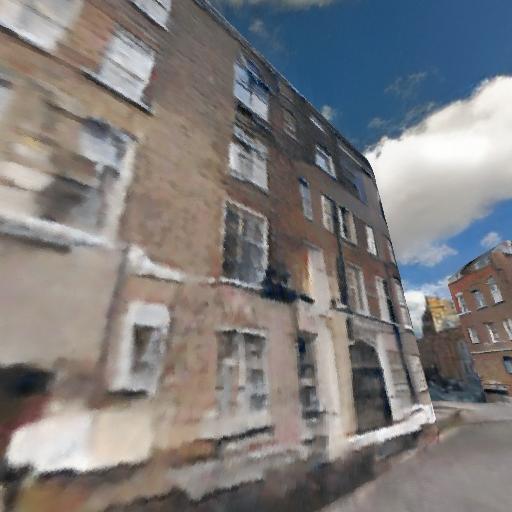} \\
    
    \includegraphics[width=\sz]{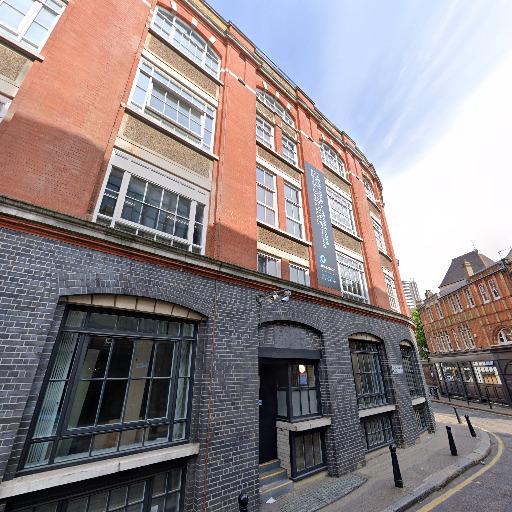} &
    \includegraphics[width=\sz]{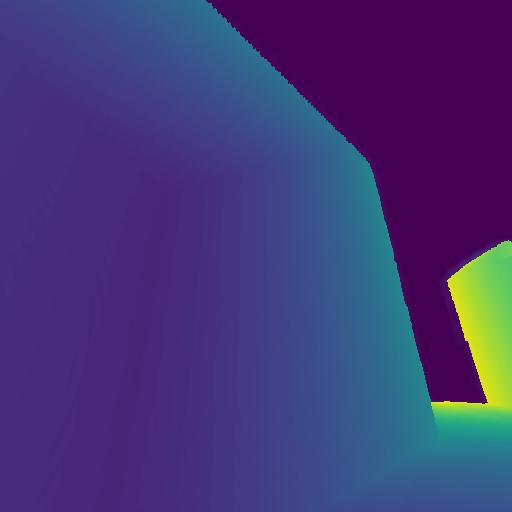} &
    \includegraphics[width=\sz]{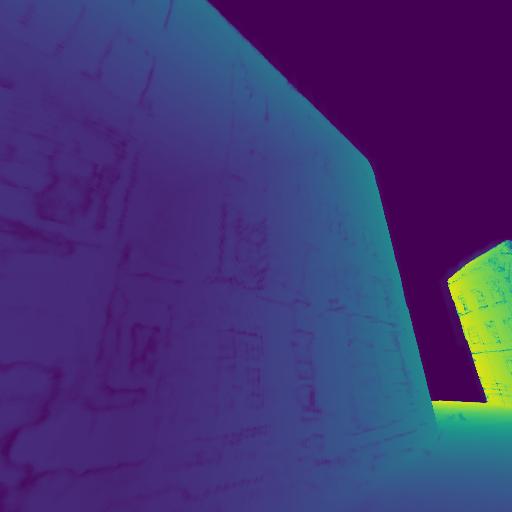} &
    \includegraphics[width=\sz]{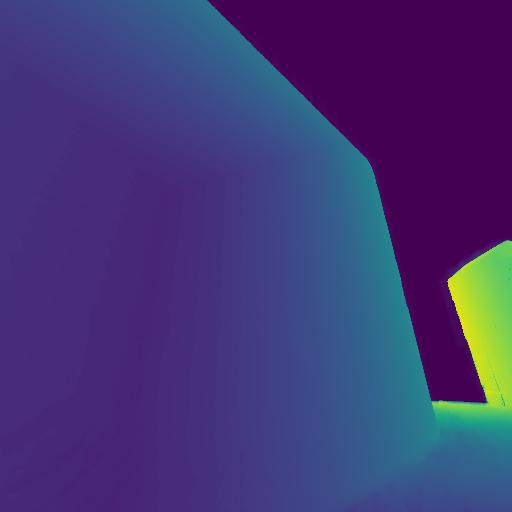} \\
    
    \end{tabular}
    
    \caption{\textbf{Additional qualitative ablation study.} We present further qualitative results for various ablations of our method. The rendered images visibly contain more details and the depths are recovered better with our full method. \textbf{The second line of each example shows the depth in pseudo colors, except the bottom left ones which are GT images.}}
    \label{fig:ablation_supp}
\end{figure}

\boldparagraph{Supplementary experiments} including additional qualitative baseline comparisons and a qualitative ablation study, are presented in \cref{fig:sota_supp} and \cref{fig:ablation_supp}, respectively.
These figures serve as extensions of \cref{fig:sota} and \cref{fig:ablation}.

\boldparagraph{The inferior consistency of MVDiffusion~\cite{mvdiffusion}} compared to its original paper is evident.
In addition to the overlap ratio mentioned in \cref{sec:qualitative}, we believe this may also be attributed to factors such as text prompt and depth detail.
MVDiffusion~\cite{mvdiffusion} utilizes long and diversified text prompts to provide the network with rich scene information. 
To ensure a fair comparison with our approach and other baselines, we use a universal text prompt, which may increase the difficulty of learning color information or detailed texture. 
As for the depth detail, the indoor depths in MVDiffusion~\cite{mvdiffusion} exhibit high quality, even allowing for the recognition and inference of object boundaries. 
In contrast, the outdoor depths in HoliCity~\cite{holicity} appear more ``flat'' and lack high-frequency details such as windows and facade decorations. 
Thus, learning the mapping from geometry to appearance in the HoliCity dataset becomes more challenging compared to the indoor scenarios in MVDiffusion~\cite{mvdiffusion}.

\section{Limitations and discussion}
\label{sec:supp_lim}

Although Sat2Scene produces photo-realistic street-view videos with robust temporal consistency and outperforms existing methods, there are still several major limitations.

\boldparagraph{Large-scale generation}.
Our model may not handle very large-scale scenes, \eg, a city-scale scene due to potentially limited computation resources.
Generating blocks separately can be an alternative to solve the scale problem, but it also introduces the potential problem of texture discontinuity between neighboring blocks.

\boldparagraph{Computation only on surfaces}.
Our model only performs on the potentially visible surfaces rather than all the surfaces of the scene to reduce computational efforts.
The invisible grounds at the base of the buildings were also removed.

\boldparagraph{Style diversity}.
The generated building textures do not hold very well diversity, which can be attributed to several factors:
\textbf{(1)} The GT geometry in HoliCity~\cite{holicity} dataset is not sufficiently detailed, lacking representations of subtle features such as slightly elevated sidewalks or recessed windows on building facades.
Such a limitation can hinder the learning process for mapping geometry to appearance.
\textbf{(2)} Our approach generates entire scenes simultaneously, necessitating the learning of appearance correlations among different building instances, in addition to individual building's appearance.
This may lead the network to pay less attention to diversity.
\textbf{(3)} 3D networks typically have a smaller network scale (number of layers, channels, \etc) than 2D ones due to 3D convolution, which limits their capacity to generate diverse appearances. 
As potential solutions, one can consider introducing latent diffusion models in 3D sparse space or adopting generation based on individual building instances. 
Also, incorporating semantics as input conditions may contribute to enhancing diversity.

\boldparagraph{Road surface}.
The road surface is not very well generated and there are three potential reasons stemming from the dataset:
\textbf{(1)} The presence of cars, passengers, and shadows on roads in GT images poses a challenge to the segmentation model in filtering them out, potentially resulting in black spots on the road surface during inference. 
\textbf{(2)} Due to the GT poses often looking a bit up to the sky, a small area around the camera origin tends to lack sufficient supervision signals during training. 
\textbf{(3)} The pixel ratio for buildings ($\sim$43.2\%) surpasses that of roads ($\sim$15.8\%) in GT images, leading the network to prioritize learning building facades in this unbalanced scenario. 
A potential solution for addressing (3) is to introduce a balance factor in the loss functions or employ distinct networks dedicated to handling buildings and roads, respectively.

\boldparagraph{Future direction}.
Below we discuss potential future development directions.
Instead of the current setting which is generating the whole scene, we can further divide the whole geometry into small elements, \eg, building instances, and generate textures individually, which could lead to better diversity.
In terms of network architecture, we may also follow LDMs~\cite{ldm} to perform the diffusion models in a latent space, given that there is enough 3D urban scene data to train a good auto-encoder with a sparse setting.
Furthermore, we can incorporate satellite image information mainly for road surface generation.
Also, our model can be combined with a natural language setting, or can even build a side branch similar to ControlNet~\cite{controlnet} on top of the trained one, to further extend the model to a conditional generation, \eg, having semantic information as input.

\end{document}